%% file: TPAMI.tex
\documentclass[9.5pt,journal,compsoc]{IEEEtran}
\usepackage{amsmath,amsfonts}
\usepackage{algorithm}
\usepackage{algorithmicx}
\usepackage{array}
\usepackage[caption=false,font=normalsize,labelfont=sf,textfont=sf]{subfig}
\usepackage{amssymb}
\usepackage{booktabs}
\usepackage{xcolor}
\usepackage{algpseudocode}
\usepackage{multirow, tabularx}
\usepackage{color}
\usepackage{textcomp}
\usepackage{amsmath}
\newcommand{\eg}{\textit{e.g.}~}
\newcommand{\etc}{\textit{etc.}~}
\usepackage{stfloats}
\usepackage{url}
\usepackage{verbatim}
\usepackage{graphicx}
\usepackage{cite}
\usepackage{marvosym}
\usepackage[pagebackref,breaklinks,colorlinks]{hyperref}
\usepackage[capitalize]{cleveref}
\begin{document}

\title{MovieChat+: Question-aware Sparse Memory for Long Video Question Answering}

\author{Enxin Song*, Wenhao Chai*, Tian Ye, Jenq-Neng Hwang,~\IEEEmembership{Life Fellow,~IEEE,} Xi Li,~\IEEEmembership{Senior Member,~IEEE,} Gaoang Wang,~\IEEEmembership{Member,~IEEE}
\IEEEcompsocitemizethanks{
\IEEEcompsocthanksitem * Equal Contributions.
\IEEEcompsocthanksitem E. Song and G. Wang are with Zhejiang University-University of Illinois at Urbana-Champaign Institute, Zhejiang University, China.
\IEEEcompsocthanksitem W. Chai and J.-N. Hwang are with the Department of Electrical and Computer Engineering, the University of Washington, USA.
\IEEEcompsocthanksitem T. Ye is with Robotics and Autonomous Systems Thrust, Hong Kong University of Science and Technology (GuangZhou), China.
\IEEEcompsocthanksitem X. Li is with the College of Computer Science and Technology, Zhejiang University, China.
}
}



\IEEEtitleabstractindextext{%
\input{text_tpami/0_abs}}
\maketitle

\input{text_tpami/1_intro}
\input{text_tpami/2_related}
\input{text_tpami/3_method}

\input{text_tpami/4_data}

\input{text_tpami/5_experiment}
\input{text_tpami/6_conclusion}
\input{text_tpami/7_acknow}


{\small
\bibliographystyle{ieee_fullname}
\bibliography{main}
}

\newpage

\input{text_tpami/8_biography}

\end{document}

%% file: text_tpami/0_abs.tex
\begin{abstract}
Recently, integrating video foundation models and large language models to build a video understanding system can overcome the limitations of specific pre-defined vision tasks. 
Yet, existing methods either employ complex spatial-temporal modules or rely heavily on additional perception models to extract temporal features for video understanding, and they only perform well on short videos. For long videos, the computational complexity and memory costs associated with long-term temporal connections are significantly increased, posing additional challenges.
Taking advantage of the Atkinson-Shiffrin memory model, with tokens in Transformers being employed as the carriers of memory in combination with our specially designed memory mechanism,  we propose MovieChat to overcome these challenges. 
We lift pre-trained multi-modal large language models for understanding long videos without incorporating additional trainable temporal modules, employing a zero-shot approach. 
Additionally, in our new version, MovieChat+, we design an enhanced vision-question matching-based memory consolidation mechanism to more significantly anchor the predictions of the visual language models in the relevant visual content.
MovieChat achieves state-of-the-art performance in long video understanding, along with the released MovieChat-1K benchmark with 1K long video, 2K temporal grounding labels, and 14K manual annotations for validation of the effectiveness of our method. The code along with the dataset can be accessed via the following \href{https://github.com/rese1f/MovieChat}{link}.
\end{abstract}


%% file: text_tpami/1_intro.tex
\section{Introduction}
Recent advances in Large Language Models (LLMs)~\cite{gpt4,brown2020language,touvron2023llama,chiang2023vicuna,taori2023stanford} achieve great success in Natural Language Processing (NLP). It is a natural progression to introduce multi-modality~\cite{chai2022deep} into LLMs and turn it into Multi-modal Large Language Models (MLLMs), which are able to conduct multimodal rationalization and understanding. MLLMs have shown incredible emergent capabilities in various multimodal tasks such as perception (\eg, existence, count, position, OCR)~\cite{wang2023visionllm, maaz2023video, alayrac2022flamingo,zhu2023minigpt,li2023otter,li2023blip}, commonsense reasoning~\cite{gao2023llama,zhu2023minigpt,li2023otter,li2022blip,li2023blip,gong2023multimodal,su2023pandagpt, maaz2023video}, embodied agent~\cite{zhao2023see,zhao2024hierarchical,zhao2024we}, and code reasoning~\cite{fu2023mme,lyu2023macaw,ye2023mplug,dai2023instructblip,liu2023visual,gao2023llama}, resulting in a potential path to Artificial General Intelligence (AGI). Compared to LLMs and other task-specific models, MLLMs provide a more human-like interpretation of the scenarios, a user-friendly interface for interaction, and a broader range of capabilities.

\input{fig_tpami/intro}
\input{fig_tpami/compare}

Existing vision-centric MLLMs follow the paradigm that utilizes pre-trained LLMs and visual encoders with additional learnable modules (Q-former~\cite{dai2023instructblip,li2022blip,li2023blip,zhang2023video} or simple projection layer~\cite{driess2023palm,maaz2023video,liu2023visual,su2023pandagpt}). In the video understanding field, some previous works~\cite{zhang2023video,maaz2023video} follow this paradigm to build video MLLMs, while works in the other paradigm~\cite{wang2023chatvideo,li2023videochat} combine existing visual perception tools (\eg, tracking and classification) and LLMs through application programming interface to build a system without training. 
However, these methods either employ complex spatial-temporal modules or heavily rely on additional perception tools to acquire temporal information for video understanding. 
Additionally, when handling long videos, the computational complexity and memory costs associated with long-term temporal connections are significantly increased, posing additional challenges, as shown in Fig.~\ref{fig:intro}. Furthermore, there is also a lack of a standardized benchmark to evaluate the capabilities of these systems.

To the best of our knowledge, our work is the first to address long video understanding tasks ($\textgreater 10$K frames). We argue that the computational complexity, memory costs, and long-term temporal connection are the main challenges for understanding long videos. Unlike other approaches that require training additional temporal modules~\cite{zhang2023video,maaz2023video} in Fig.~\ref{fig:compare}, our MovieChat lifts pre-trained MLLMs to understand long videos without the need for additional trainable temporal modules, employing a zero-shot approach.
Inspired by the Atkinson-Shiffrin memory model~\cite{atkinson1968chapter}, we propose a memory mechanism to address long video understanding tasks. This mechanism comprises a rapidly updated short-term memory and a compact, thus, sustained long-term memory. 
In our updated version, namely \textbf{MovieChat+}, we design a vision-question matching-based memory consolidation mechanism to enhance the compactness of memory. This mechanism significantly anchors the predictions of the visual language models in the relevant visual content. 
Our MovieChat+ significantly improves upon the initial version and outperforms the state-of-the-art in both short and long video question-answering tasks, surpassing even methods specifically tailored for short video question-answering challenges.
As shown in Fig.~\ref{fig:intro}, our approach outperforms other existing methods in terms of Video Random-Access Memory (VRAM) cost. We also release a new benchmark, MovieChat-1K, with 1K long videos and 14K manual question-answering pairs for validation of the effectiveness of our proposed method. Additionally, we expand MovieChat-1K with 2K temporal labels. 
The contributions of this work are summarized as follows:
\begin{itemize}
    \item We present MovieChat, the first framework designed to support long-term videos ($\textgreater 10$K frames), leveraging pre-trained MLLMs and employing a zero-shot, training-free memory consolidation mechanism.
    \item Our updated version, \textbf{MovieChat+}, enhances memory compactness through the implementation of a vision-question matching-based memory consolidation technique. This improvement significantly surpasses the initial version and outperforms the current state-of-the-art in both short and long video question-answering tasks.
    \item We have released the first long-video understanding benchmark, MovieChat-1K, which now includes an expansion to 2K temporal labels compared to the initial version. We conducted extensive quantitative evaluations and case studies to assess the comparable performance of both understanding capability and inference cost.
\end{itemize}

%% file: fig_tpami/intro.tex
\begin{figure}[t]
    \centering
    \includegraphics[width=1\linewidth]{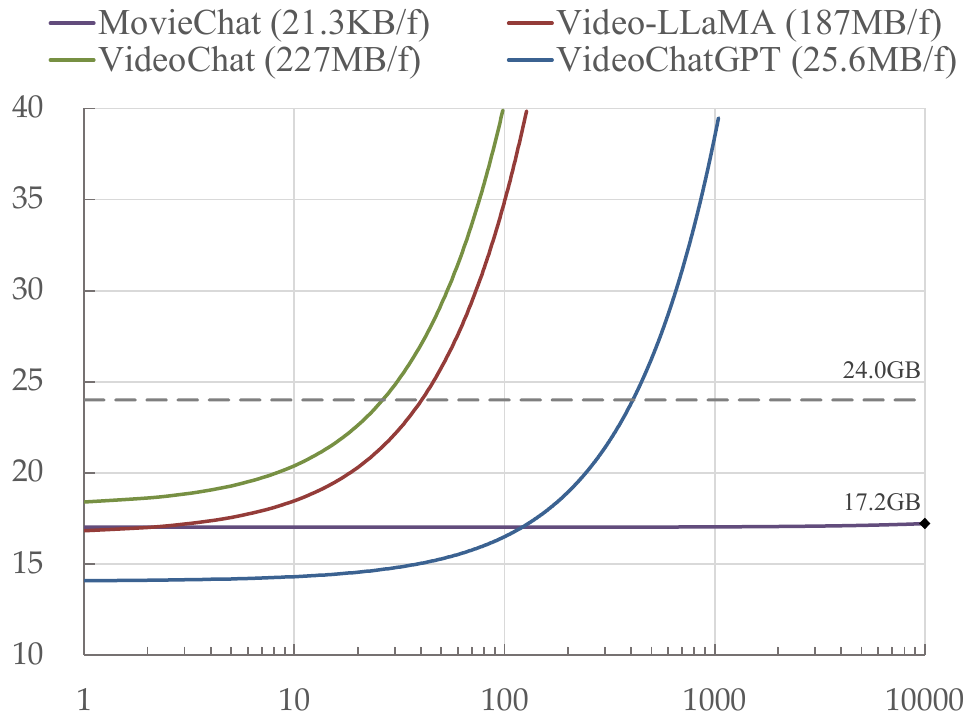}
    \caption{Video random-access memory (VRAM) cost under gigabyte (GB) (y-axis) v.s. frame number (x-axis) comparison. We test the visual-only inference of all methods at a resolution of $224 \times 224$ without frame sampling. While the previous method can only support around $100$ frames of inference, MovieChat can handle videos with $\textgreater 10$K frames on a 24GB graphics card. MovieChat has a $10000 \times$ advantage over other methods in terms of the average increase in VRAM cost per frame ($21.3$KB to $\sim200$MB per frame).}
    \label{fig:intro}
\end{figure}

%% file: fig_tpami/compare.tex
\begin{figure*}[t]
    \centering
    \includegraphics[width=\linewidth]{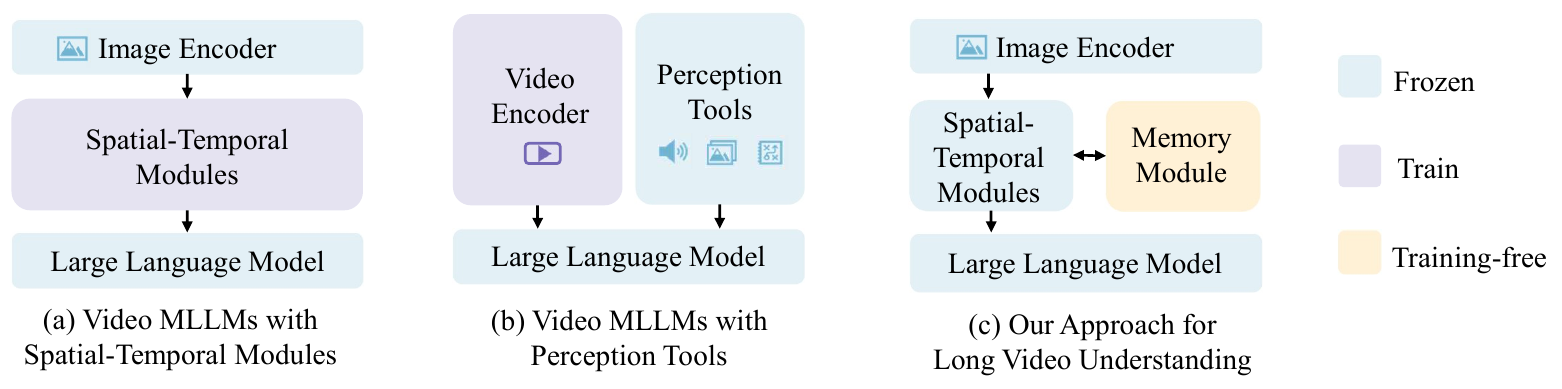}
    \caption{Comparison between existing video MLLMs and our appraoch. While previous works either employ complex spatial-temporal modules or heavily rely on additional perception tools, and struggle with long videos, our approach is the first to address long video understanding tasks without training.}
    \label{fig:compare}
\end{figure*}

%% file: text_tpami/2_related.tex
\section{Related Works}

\subsection{Multi-modal Large Language Models}
LLMs~\cite{gpt4,brown2020language,touvron2023llama,touvron2023llama2,chiang2023vicuna,taori2023stanford} have achieved great success in natural language processing (NLP) tasks recently. Many works try to build MLLMs~\cite{alayrac2022flamingo,zhu2023minigpt,li2023otter,li2023blip,gong2023multimodal,ye2023mplug,dai2023instructblip,wang2023visionllm,maaz2023video,gao2023llama, weng2024longvlm, ma2023vista} by combining models of other modalities.
Flamingo~\cite{alayrac2022flamingo} bridges powerful pre-trained vision-only and language-only models and achieves state-of-the-art performance with few-shot learning.
BLIP-2~\cite{li2023blip} proposes a generic and efficient pre-training strategy that bootstraps vision-language pre-training from an off-the-shelf frozen pre-trained image encoder and a frozen large language model.
MiniGPT-4~\cite{zhu2023minigpt} also aligns a frozen visual encoder with a frozen LLM, Vicuna~\cite{chiang2023vicuna}, using just one projection layer to realize the system.
Otter~\cite{li2023otter} showcases improved instruction-following ability and in-context learning.
In the video field, ChatVideo~\cite{wang2023chatvideo} treats tracklets as the basic video unit and allows users to interact with the LLMs.
VideoChat~\cite{li2023videochat} integrates video foundation models and LLMs via a learnable neural interface, excelling in spatiotemporal reasoning, event localization, and causal relationship inference. 
Video-LLaMA~\cite{zhang2023video} further leverages pre-trained models ImageBind~\cite{girdhar2023imagebind} and LLaMA~\cite{touvron2023llama}, bootstraping cross-modal training in videos following BLIP-2.
Yet, these methods fail to handle long video understanding because of high computation complexity, large memory cost, and weak long-term temporal connection. Therefore, our main effort is to introduce an effective memory mechanism to overcome these challenges.

\input{fig_tpami/overview}

\subsection{Long Video Understanding}

Understanding long videos is a challenging task in computer vision. Prior arts use 3D CNN for long-term feature bank~\cite{wu2019long}, object/human-centric motion~\cite{wu2021towards,rohrbach2017generating}, or other forms~\cite{wu2022memvit,sener2020temporal} as video representations. MIST~\cite{gao2023mist} decomposes dense self-attention into a cascade segment and
region selection module to increase the computation efficiency for understanding minutes of long videos. Building long-form video understanding datasets is challenging and rarely explored.  \cite{shou2021generic} captures large scale data from Kinetics-400~\cite{carreira2017quo}, but only for generic event boundary detection tasks. \cite{soldan2022mad} creates a language grounding benchmark from audio descriptions of movies, but it lacks long-term understanding evaluation. \cite{tapaswi2016movieqa} successfully builds a benchmark contains multiple sources of information (\eg, video clips, plots, and DVS) for question-answering tasks in the movie field. There are also several datasets of video-caption/description pairs among various domains, such as cooking (\eg, MPII Cooking~\cite{rohrbach2012database,rohrbach2012script,rohrbach2016recognizing} and TACoS~\cite{regneri2013grounding,rohrbach2014coherent}), instruction (\eg, HowTo100M~\cite{miech2019howto100m} and HiREST~\cite{zala2023hierarchical}), Ego~\cite{mangalam2023egoschema}, and movie (\eg, MovieQA~\cite{tapaswi2016movieqa} and MovieNet~\cite{huang2020movienet}) from different sources such as YouTube~\cite{chen2011collecting,zeng2016title,miech2019howto100m}, Twitter~\cite{awad2017trecvid,awad2018trecvid,awad2020trecvid,awad2021trecvid}, and Internet~\cite{bain2021frozen}. 
Yet, those datasets lack diverse and fine-grained dense captioning for long videos.

\subsection{Memory Models in Vision Tasks}

There are some prior works exploring memory models~\cite{squire2015memory} in various vision tasks in videos, such as video object segmentation~(VOS)~\cite{cheng2022xmem,hu2021learning,seong2020kernelized,seong2021hierarchical}, multi-object tracking~(MOT)~\cite{cai2022memot,hao2022umotma,xin2022multi,allen2006multiple}, visual object tracking~(VOT)~\cite{zhou2023memory,yang2018learning,liu2017mavot,ma2018adaptive}, and action understanding~\cite{wang2023memory}.
MeMOT~\cite{cai2022memot} builds a large spatiotemporal memory that stores the past observations of the
tracked objects. 
XMem~\cite{cheng2022xmem} develops an architecture that incorporates multiple independent yet deeply connected feature memory storage to handle long videos with thousands of frames. Recently, \cite{zhou2024streaming} propose a memory streaming method to support dense video captioning.
We learn from the experience of those prior arts and further adopt an effective memory mechanism in combination with LLMs. 

Our method focuses on reducing the redundancy of visual tokens in the video and building a memory mechanism to pass the information among a large temporal range.

%% file: fig_tpami/overview.tex
\begin{figure*}[t]
    \centering
    \includegraphics[width=0.95\linewidth]{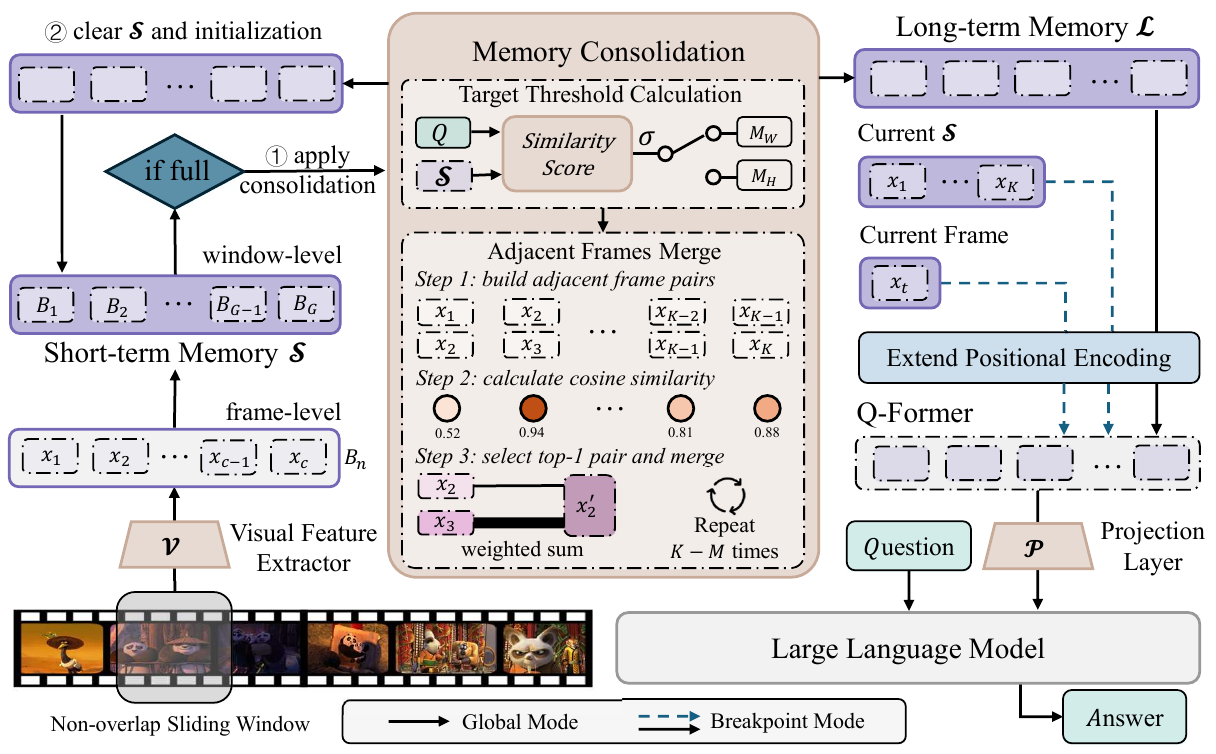}
    \caption{\textbf{Illustration of MovieChat, a training-free framework with question-aware memory consolidation mechanism.} MovieChat extracts video features with a sliding window and represents them in token form, which are then sequentially fed into the short-term memory frame by frame. When the fixed-length short-term memory reaches its preset limit, the earliest tokens are popped and consolidated into the long-term memory. Our approach incorporates two distinct inference modes: the global mode, which exclusively utilizes the long-term memory, and the breakpoint mode, which additionally incorporates the current short-term memory as part of the video representation. The breakpoint mode allows for understanding the video at a specific moment in time. After passing through a projection layer, the video representation is inputted into a large language model for interaction with the user.}
    \label{fig:overview}
\end{figure*}


%% file: text_tpami/3_method.tex
\section{MovieChat}

\input{alg/alg}

\subsection{Overview}
Our proposed method, MovieChat, comprises several key components, including the frame-wise visual feature extractor, the short-term memory module, the long-term memory module with question-aware consolidation strategy (our improved version, namely MovieChat+), the video projection layer, and the Large Language Model (LLM), as illustrated in Fig.~\ref{fig:overview}. our approach is designed for ultra-long videos ($\textgreater 10$K frames) understanding through interactive dialogue with the user. To address the impractical storage demands of concurrently storing a vast number of frames in both GPU memory and RAM, we employ a sliding window approach to efficiently process the video. The short-term memory module embeds dense tokens with sliding windows, and the long-term memory module periodically updates based on question-aware consolidation. MovieChat supports two inference modes: breakpoint mode and global mode. Breakpoint mode is used to understand a specific moment in the video, providing insights and answers based on that particular frame or scene. Global mode, on the other hand, is employed to comprehend the entire video as a whole, enabling a comprehensive understanding of the overall content and context.

\subsection{Visual Feature Extraction}
For visual feature extraction, instead of utilizing video-based foundational models~\cite{arnab2021vivit,liu2022video}, we simply use an image-based model to get frame-wise features in the form of tokens. To be specific, we utilize pre-trained models as our visual feature extractor, including the ViT-G/14 from EVA-CLIP~\cite{fang2022eva} and the Q-former from BLIP-2~\cite{li2023blip2}. This is mainly because 1) there are few video foundation models that make good alignment with text, and 2) our proposed memory mechanism can effectively capture temporal features. Given a raw video, the visual input $\mathbf{v} \in \mathbb{R}^{T \times 3 \times H \times W}$ is a sequence of $T$ RGB frames of size $H \times W$ sampled from the video. The visual features are extracted in a sliding window manner:
\begin{equation}
    B_{n} = \{ \mathbf{x}_{i}= \mathcal{V}(\mathbf{v}_{i}) \mid \forall i = 1,...,C\}, n = 1,...,\lceil \frac{T}{C} \rceil,
\end{equation}
where $B_{n}$ is the $n$-th video clip feature within the sliding window spanning $C$ frames. $\mathcal{V}(\cdot)$ is the visual feature extractor, taking as input a single frame {$\mathbf{v}_{i}\in \mathbb{R}^{3 \times H \times W}$}. {$\mathbf{x}_{i}$} $\in \mathbb{R}^{N \times D}$ denotes $N$ extracted visual tokens with respect to each frame, and $D$ is the feature dimension of each token.

\input{fig_tpami/question_aware}

\subsection{Short-term Memory}
The short-term memory maintains a fixed-length buffer to temporarily hold frame tokens. The previously extracted visual features by sliding window $G$ times without further processing are used to construct short-term memory, which can be described by the following formulation:
\begin{equation}
    \mathcal{S} = \bigcup_{n}{B}_{n} = \{ \mathbf{{x}}_{i} \mid \forall i = 1, ..., K\}, n=1,..,G,
\end{equation}
where $\mathcal{S}$ is the short-term memory, and $K$ is equal to $ C \times G$. Note that we configure short-term memory to contain a fixed length of $K$ frames. The design stems from the fundamental purpose of short-term memory, which is to aid in the interpretation and understanding of video content by leveraging contextual information from recent, short-term segments of the video.

As a new batch of visual tokens enters, when the short-term memory reaches its capacity, we pop the currently stored frames to the memory consolidation module and clear the short-term memory. The output video feature obtained from the consolidation module augments the long-term memory; on the other hand, it re-initializes the short-term memory with this feature. The initialization aims to communicate information between different sliding windows, thereby achieving more efficient compression.

\input{fig_tpami/benchmark}

\subsection{Question-aware Long-term Memory (MovieChat+)}

The long-term memory can effectively avoid the problem of catastrophic knowledge forgetting, which is crucial for handling long video understanding tasks. The features stored in short-term memory are dense tokens, but due to the limitations of GPU memory and computation cost, storing all the tokens dropped from short-term memory into long-term memory buffer in sequence is infeasible. Besides, we observe significant temporal redundancy in videos, where activities span multiple frames with minimal visual changes. Additionally, only a small fraction of the entire long video content is relevant to the given question in practice. To this end, in our updated version, MovieChat+, we propose to merge adjacent frames based on their relevance to specific questions, thereby streamlining video feature representation and enhancing encoding efficiency. This method converts dense tokens into sparse memories centered on pertinent questions, which are stored in long-term memory.

To be specific, as shown in Algorithm~\ref{alg:alg}, we first utilize a pre-trained text encoder $\mathcal{T}(\cdot)$ to encode the specific question $\mathbf{Q}$ to the same embedding space as the visual features, which can be formulated as:
\begin{equation}
    \mathbf{q} = \mathcal{T}(\mathbf{Q}).
\end{equation}

We then calculate the average cosine similarity $s^{q}$ between each frame feature $\mathbf{x}_{i}$ within the short-term memory and the encoded question $\mathbf{q}$, which can be formulated as:
\begin{equation}
    s^{q} = \frac{1}{K} \sum_{i=1}^{K} \left [ \cos(\mathbf{x}_{i}, \mathbf{q}) \right ].
\end{equation}

When we watch a video with a question in mind, we tend to skip over largely irrelevant segments. This is the motivation for using the calculated similarity between questions and visual features. If the visual features in short-term memory are highly related to the questions, we merge fewer into long-term memory; otherwise, we merge more during consolidation. We consider the target merging threshold $M$, \textit{i.e.}, the number of consolidated frames after merging using the similarity calculated above.

We compare the average similarity $s^{q}$ to the threshold $\sigma$ to assess its relevance to the question.
The target merging threshold $M$ is formulated as follows,
\begin{equation}
    M = 
    \begin{cases} 
    M_{0} & \text{if } s^{q} \textgreater \sigma \\
    \alpha M_{0} & \text{otherwise}
    \end{cases}
\end{equation}
For segments with high relevance, we set a base merging threshold $M = M_{0}$. 
For segments with low relevance, we set a compression coefficient $\alpha$, reducing the target merging threshold to $M = \alpha M_{0}$, thus compressing fixed-length segments into fewer ones.

Following the methodology outlined in ToMe~\cite{bolya2022token}, we then periodically perform memory consolidation by merging the most similar tokens in the adjacent frames.
Here, we calculate the average cosine similarity $s^{f}$ among $N$ embedded tokens, as the tokens can effectively summarize the information of each frame:
\begin{equation}
    s^{f} = \frac{1}{N} \sum_{j=1}^{N} \left [ \cos(\mathbf{x}_{i}^{j}, \mathbf{x}_{i+1}^{j}) \right ].
\end{equation}

Our goal is to keep $M$ frames after every merge operation, which also embeds rich information stored in the long-term memory. $M$ is the hyper-parameter to control the trade-offs between performance and efficiency. Therefore, we greedily merge each set of adjacent frames with the highest similarity based on hierarchical clustering via weighted averaging, as shown in Algorithm~\ref{alg:alg}. The merge operation is iteratively conducted until the token count reaches the predefined value set $M$ for each consolidation operation, resulting in the output video feature $\mathbf{v'} \in \mathbb{Z}^{M \times 3 \times H \times W}$. Consequently, the dense tokens of the entire video are compressed to varying degrees based on their similarity to the question and are stored within the long-term memory as shown in Fig.~\ref{fig_tpami:ques_aware}. The above algorithm is parameter-free and can be easily plugged into a frame-based video encoder. Although the frame similarity calculation brings additional computing overhead, it is negligible compared to the efficiency gained by reducing stored frames.

\textbf{Extend Positional Encoding.} For long-term memory, the number of tokens exceeds the maximum length of the positional encoding from the pre-trained model. Thus, our model utilizes the positional encoding mechanism following BERT~\cite{kenton2019bert}, which results in a portion exceeding the length threshold $n$ without available positional encoding. In order to handle long enough memory, we adopt the hierarchically decomposed positional encoding method proposed by Su~\cite{pos}, which allows us to extend the absolute positional encoding of length from $n$ to $n^2$.

\subsection{Inference}

Previous methods always use the representation of the whole video to conduct understanding and question-answering. While this method provides a broad overview, it often struggles with accurately localizing specific moments or details in long videos. To this end, we propose two inference modes, global and breakpoint, for long video understanding tasks as follows.

\input{fig_tpami/certi}

\noindent \textbf{Global Mode.}
Global mode is defined as the understanding and question-answering for the whole video. Under this mode, the focus is on capturing the details of the full duration of the video. Therefore, we only use long-term memory $\mathcal{L}$ as the video representation $\mathbf{V}$.

\noindent \textbf{Breakpoint Mode.}
Breakpoint mode is defined as understanding specific moments in a video. Since events possess continuity, we need to consider not only the information directly related to the moments stored in short-term memory $\mathcal{S}$ but also the information indirectly related stored in long-term memory $\mathcal{L}$. Therefore, we hypothesize that when querying the movie at a specific moment $t$, the video representation $\mathbf{V}$ should be the aggregation of $\mathcal{L}$, $\mathcal{S}$, and the current video frame feature $\mathbf{x}_{t}$. We observe that straightforward concatenation of these elements delivers outstanding results, and we defer the investigation of alternative aggregation methods to future research.


Subsequently, the video representation goes through a Q-former and a linear projection layer before being fed into the LLM $\mathcal{O}$, which can be formulated as:
\begin{equation}
    \mathbf{A}_t = \mathcal{O}(\mathbf{Q}, \mathcal{P} (\{\mathcal{L}, \mathcal{S}, \mathbf{x}_{t}\})),
\end{equation}
where $\mathcal{P}$ is the projection from visual space to text space, $\mathbf{A}_t$ represents the answer or instruction of the breakpoint, and $\mathbf{Q}$ is employed to denote the question, respectively.

\input{tab_tpami/question_type}

%% file: alg/alg.tex
\definecolor{global}{RGB}{21,96,130}
\definecolor{breakpoint}{RGB}{51,0,111}

\algnewcommand{\Comment}[1]{\textcolor{blue}{\(\triangleright\) #1}}
\begin{algorithm}[t]
\caption{Memory consolidation}\label{alg:alg}
\begin{algorithmic}
\Require $\mathcal{S}$ \hfill{\textcolor{global}{\(\triangleright\) short-term memory}}
\vspace{-10pt}
\State \For{$\mathbf{x}_{i}$ in $\mathcal{S}$}
\State \hspace{0.5cm} $s^{q} \gets sim(\mathbf{x}_{i}, \mathbf{q})$ \hfill\textcolor{global}{\(\triangleright\) question and frame similarity}
\EndFor
\vspace{-10pt}
\State \If {\( \text{mean}(s^{q}) > \sigma \)} \hfill \textcolor{global}{\(\triangleright\) target merging threshold}
\State \hspace{0.5cm} $M \gets M_0$
\vspace{-10pt}
\State \Else
\State \hspace{0.5cm} $M \gets \alpha M_0$
\EndIf
\vspace{-10pt}
\State \While{$len(\mathcal{S}) \textgreater M$} \hfill \textcolor{global}{\(\triangleright\) iterative merge}
\vspace{-10pt}
\State \For{$\mathbf{x}_{i}$ in $\mathcal{S}$}
\State \hspace{0.5cm} $s^{f} \gets sim(\mathbf{x}_{i}, \mathbf{x}_{i+1})$ \hfill\textcolor{global}{\(\triangleright\)tokens similarity}
\EndFor
\State \hspace{0.25cm} $m \gets max(s^{f})$ \hfill\textcolor{global}{\(\triangleright\)the maximum value index}
\State \hspace{0.25cm} $\mathbf{x}_{m} \gets merge (\mathbf{x}_{m},\mathbf{x}_{m+1})$ \hfill\textcolor{global}{\(\triangleright\)merge}
\State \textbf{del} $\mathbf{x}_{m+1}$
\EndWhile
\label{alg}
\end{algorithmic}
\end{algorithm}

%% file: fig_tpami/question_aware.tex
\begin{figure}[t]
    \centering
    \includegraphics[width=\linewidth]{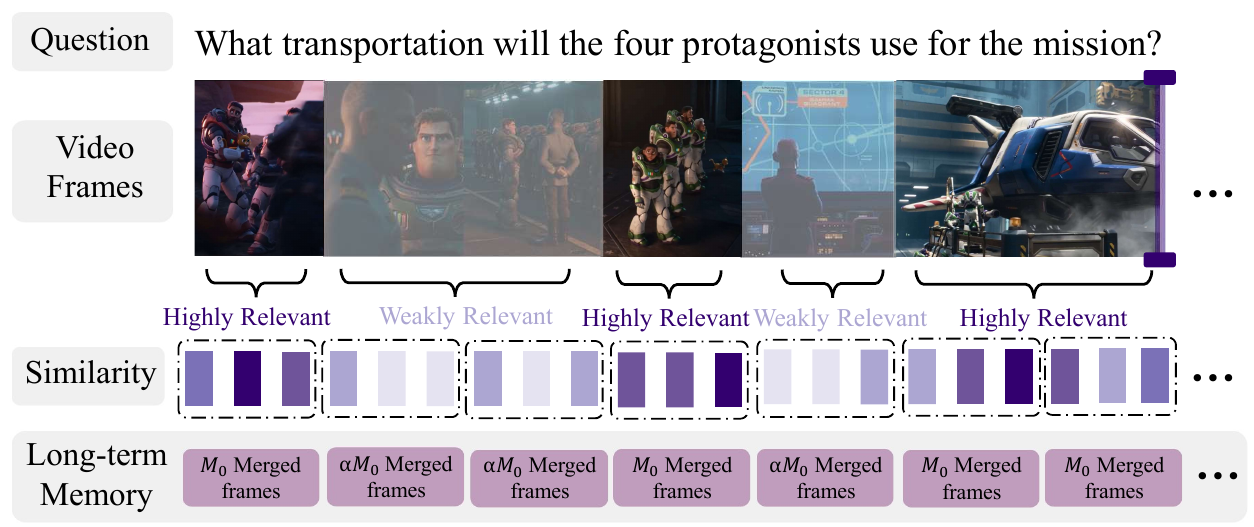}
    \captionsetup{font={scriptsize}}
    \caption{Visualization of the question-aware memory consolidation strategy. If frames in the short-term memory are highly relevant to the question (indicated by dark blocks), a sparse compression strategy is implemented, compressing the frames in the short-term memory into a relatively larger number of frames. Conversely, weakly relevant segments (indicated by light blocks) undergo a dense compression strategy, compressing the frames in the short-term memory into a much smaller number of frames.}
    \label{fig_tpami:ques_aware}
\end{figure}

%% file: fig_tpami/benchmark.tex
    

\begin{figure*}[!t]
\centering
\subfloat[]{\includegraphics[width=2.4in]{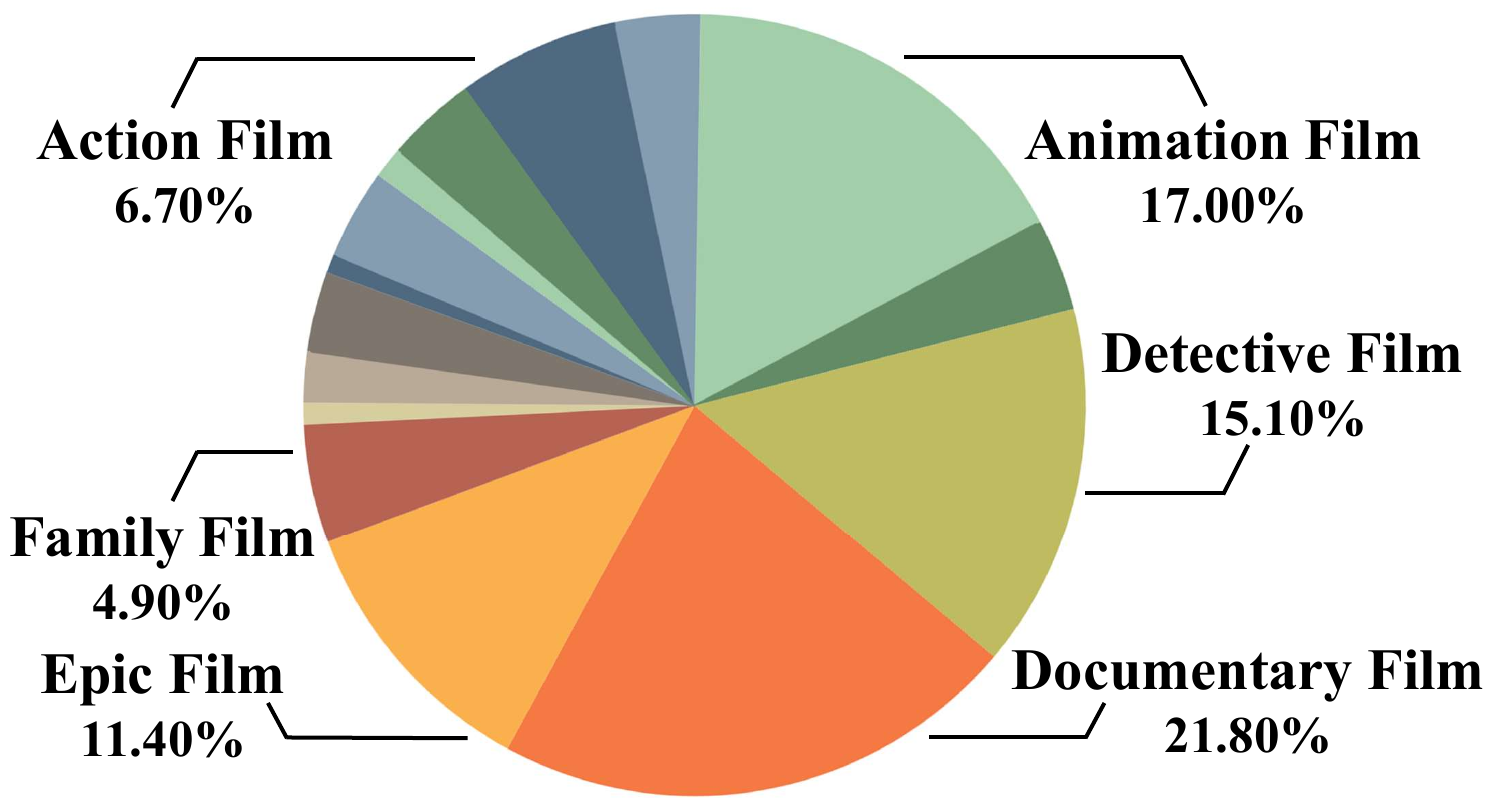}%
\label{fig_tpami:category}}
\hfil
\subfloat[]{\includegraphics[width=2.4in]{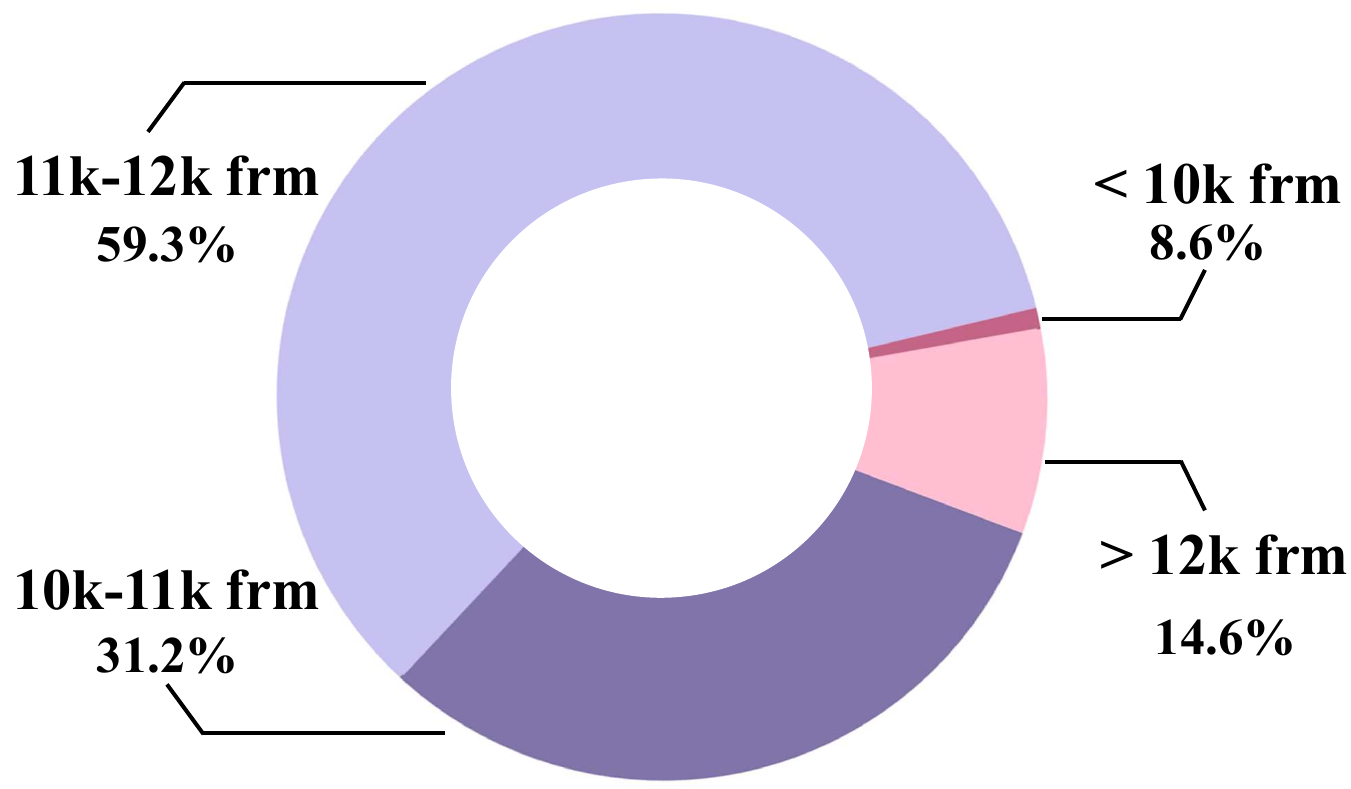}%
\label{fig_tpami:video_length}}
\hfil
\subfloat[]{\includegraphics[width=2.3in]{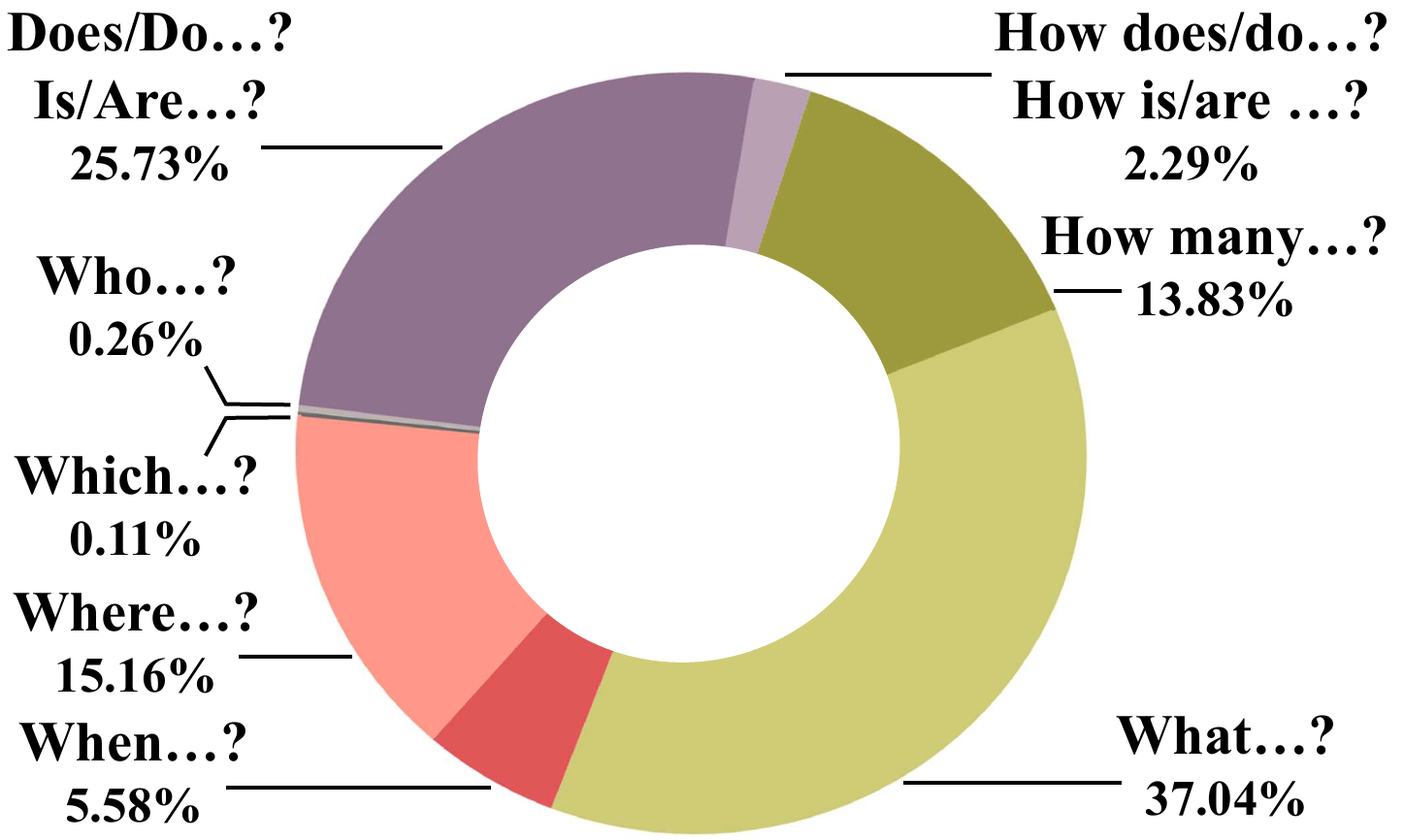}%
\label{fig_tpami:question_type}}
\caption{Video-text statistics in MovieChat-1K. It encompasses a diverse set of categories, gathered from multiple question types and containing a diverse distribution of clip durations. ``frm" represents the number of video frames.}
\label{fig_sim}
\end{figure*}

%% file: fig_tpami/certi.tex
\begin{figure}[t]
    \centering
    \includegraphics[width=\linewidth]{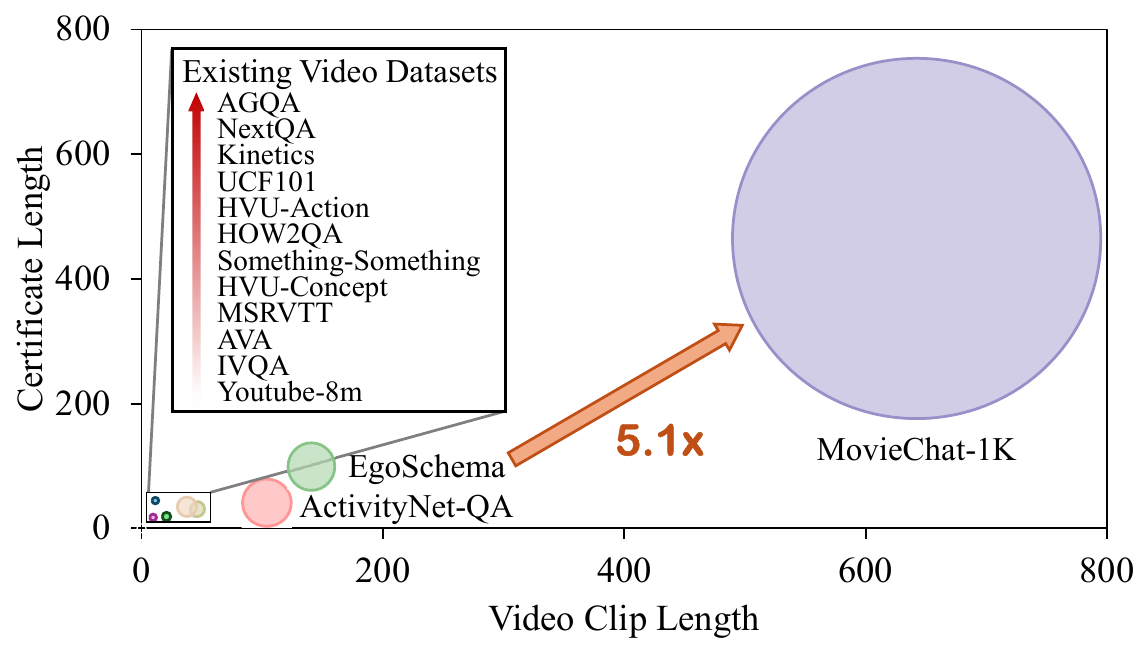}
    \vspace{-12pt}
    \captionsetup{font={scriptsize}}
    \caption{Certificate length across video datasets.}
    \label{fig_tpami:certi}
\end{figure}

%% file: tab_tpami/question_type.tex
\begin{table}[t]
\centering
\setlength{\tabcolsep}{6pt}
\renewcommand{\arraystretch}{0.9}
\captionsetup{font={scriptsize}}
\caption{Semantic Categorization of MovieChat-1K Questions}
\label{tab_tpami:q_type}
\resizebox{\linewidth}{!}{
\begin{tabular}{@{} c c | c c | c c @{}}
\toprule
 \textbf{Type} &  \textbf{Rate} & \textbf{Frame Type}&  \textbf{Text Rate} &  \textbf{Type} & \textbf{Text Rate}\\
\midrule
    Location & 21.2\% & Scene & 11.2\% & Existence & 5.0\% \\

    Time & 17.2\% & Object & 10.7\% & Weather & 3.6\% \\

    Number & 14.6\% & Action & 7.8\% & Others & 1.5\% \\
\bottomrule
\end{tabular}
}
\end{table}

%% file: text_tpami/4_data.tex
\section{A New Benchmark: MovieChat-1K}

Previous works on building long video understanding benchmarks either focus on non-question-answering tasks~(\eg, language grounding~\cite{soldan2022mad}, generic event boundary detection~\cite{shou2021generic}, user engagement and movie metadata prediction~\cite{wu2021towards}, \etc) or lack long-form understanding evaluation~\cite{huang2020movienet}. To better evaluate the performance of MovieChat, we collect a new benchmark for long video understanding tasks, MovieChat-1K, which contains 1K high-quality video clips sourced from various movies and TV series with 14K manual annotations. In our updated version, MovieChat+, we expand by an additional 2K temporal grounding labels.

\noindent \textbf{Video Source.} As shown in Fig.~\ref{fig_tpami:category}, we collect videos from 15 popular categories with varying distribution, including documentary film, detective film, animation film, etc. Among these, each video comprises multiple alternating scenes, contributing to a diverse and dynamic visual narrative within the context of the collection. We further illustrate our improved content-based categorization and analysis of MovieChat-1K questions in Tab.~\ref{tab_tpami:q_type}. The visual representation in Fig.~\ref{fig_tpami:video_length} demonstrates the clip duration distribution of MovieChat-1K. Over 90\% of the videos exhibit a duration ranging from 10K to 12K frames, while 14.6\% of videos extend beyond 12K frames. Only 8.6\% of videos have a duration of less than 10k frames. To demonstrate that MovieChat-1K is indeed a long-form dataset, we employ the same method as proposed by EgoSchema~\cite{mangalam2023egoschema} to calculate the certificate lengths. As depicted in Fig.~\ref{fig_tpami:certi}, our approach results in a certificate length that is 5.1 times longer than that of EgoSchema~\cite{mangalam2023egoschema}. Specifically, the annotated captions and questions are used as temporal tags, and the corresponding clip lengths are calculated manually. 

\input{fig_tpami/grounded}

\input{fig_tpami/segments}

\input{fig/wordcloud}

\noindent \textbf{Temporal Label Collection.} 
Following~\cite{xiao2023can}, we augment MovieChat-1K with temporal labels. Most VideoQA datasets~\cite{jang2017tgif, xu2017video} are unsuitable for exploring how to handle irrelevant redundant frames, as they are composed of short video clips (no more than 15 seconds) that have been pre-trimmed to focus solely on the pertinent content. We apply temporal labels exclusively to questions categorized under breakpoint mode. This is because questions in global mode mostly pertain to global video content (\eg, "Where does the video take place?"). Furthermore, the answers to these global mode questions are often discernible across extensive segments of the video, such as “Is there more than five different characters appearing?”. For each question-answer pair in breakpoint mode, we annotate the start and end times of the relevant segments as shown in Fig.~\ref{fig_tpami:grounded}. 

We restrict the labeling of temporal annotations in MovieChat-1K to the validation and test sets, under the premise that these labels are instrumental in assessing the ability of models to identify question-relevant video segments, rather than for training purposes. As a result, 2K question-answer pairs drawn from 200 videos are annotated with temporal labels. Fig.~\ref{fig_tpami:segments} demonstrates that most of the segments last for less than 12 seconds, with an average duration of 6.3 seconds, which is extremely short compared to the video length (approximately 700 seconds).

\input{fig/caption_length}
\input{fig/caption_wordcloud}

\noindent \textbf{Annotations Analysis.} For each video, we manually set and provide 1 dense caption for the whole video, 3 question-answering pairs for global mode, and 10 question-answering pairs with timestamps for breakpoint mode. Fig.~\ref{fig_tpami:question_type} illustrates the distribution of question types in MovieChat-1K. Note that MovieChat-1K is specifically designed for long video comprehension tasks. The majority of questions are open-ended, with only a quarter classified as multiple-choice questions, marked by initiators such as `Do,' `Does,' `Is,' or `Are.' As illustrated in  Fig.~\ref{fig:wordcloud}, we also compute the word distributions of the question-answer pairs, which includes common objects (people, clothes, etc.), time (day, night, etc.), scenes (indoor, outdoor, etc.), and so on.

To facilitate a detailed understanding of long videos, we provide a dense caption for each video. As shown in Fig.~\ref{fig:caption_length}, MovieChat-1K exhibits diverse caption lengths in the segmented clip level. Approximately two-thirds of the clips have captions with 100-149 words, while one-fifth of the clip captions have fewer than 100 words. About 11\% of clips have long captions with more than 150 words.

To analyze the word distribution of our generated
captions, we compute their distributions. The resulting word distribution is presented in Fig.~\ref{fig:caption_wordcloud}, which includes common objects (man, woman, people, girl, etc.), attributes (detective, various, small, white, etc.), locations (inside, behind, south, next, etc.), scenes (room, house, building, office, etc.), actions/events (talk, enter, leave, take, etc.), and more.

In terms of actions, MovieChat-1K captions contain nearly the same number of verbs as with the WebVid10M dataset~\cite{DBLP:journals/corr/abs-2104-00650}. To evaluate this, we use the NLTK toolkit to analyze the number of verbs in captions, focusing on extracting and tagging all unique verbs. We find a total of 109,485 verbs in the WebVid10M caption dataset, while the MovieChat-1K captions contain 102,988 unique instances of verbs. While these counts may not be entirely accurate due to our simple counting method, we believe they provide a rough indication of the actions of the two datasets.

%% file: fig_tpami/grounded.tex
\begin{figure}[t]
    \centering
    \includegraphics[width=\linewidth]{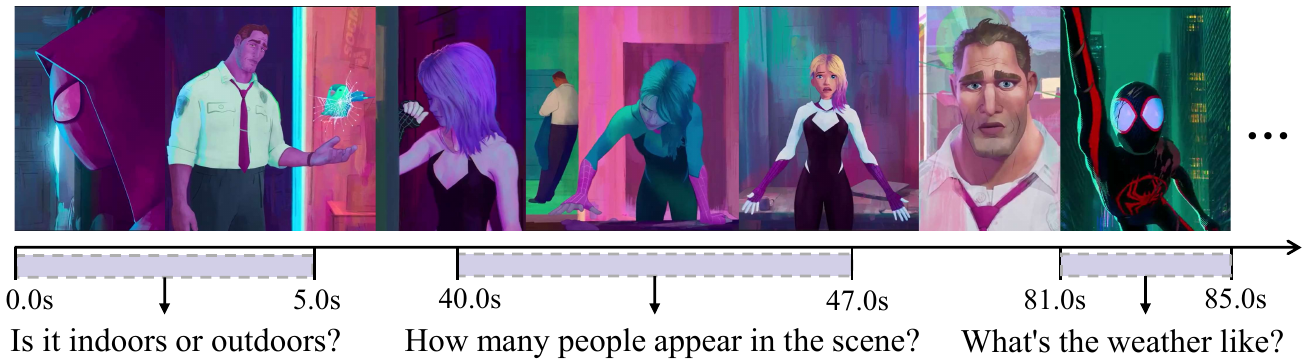}
    \captionsetup{font={scriptsize}}
    \caption{Examples of temporal labels in MovieChat-1K.}
    \label{fig_tpami:grounded}
\end{figure}

%% file: fig_tpami/segments.tex
\begin{figure}[t]
    \centering
    \includegraphics[width=\linewidth]{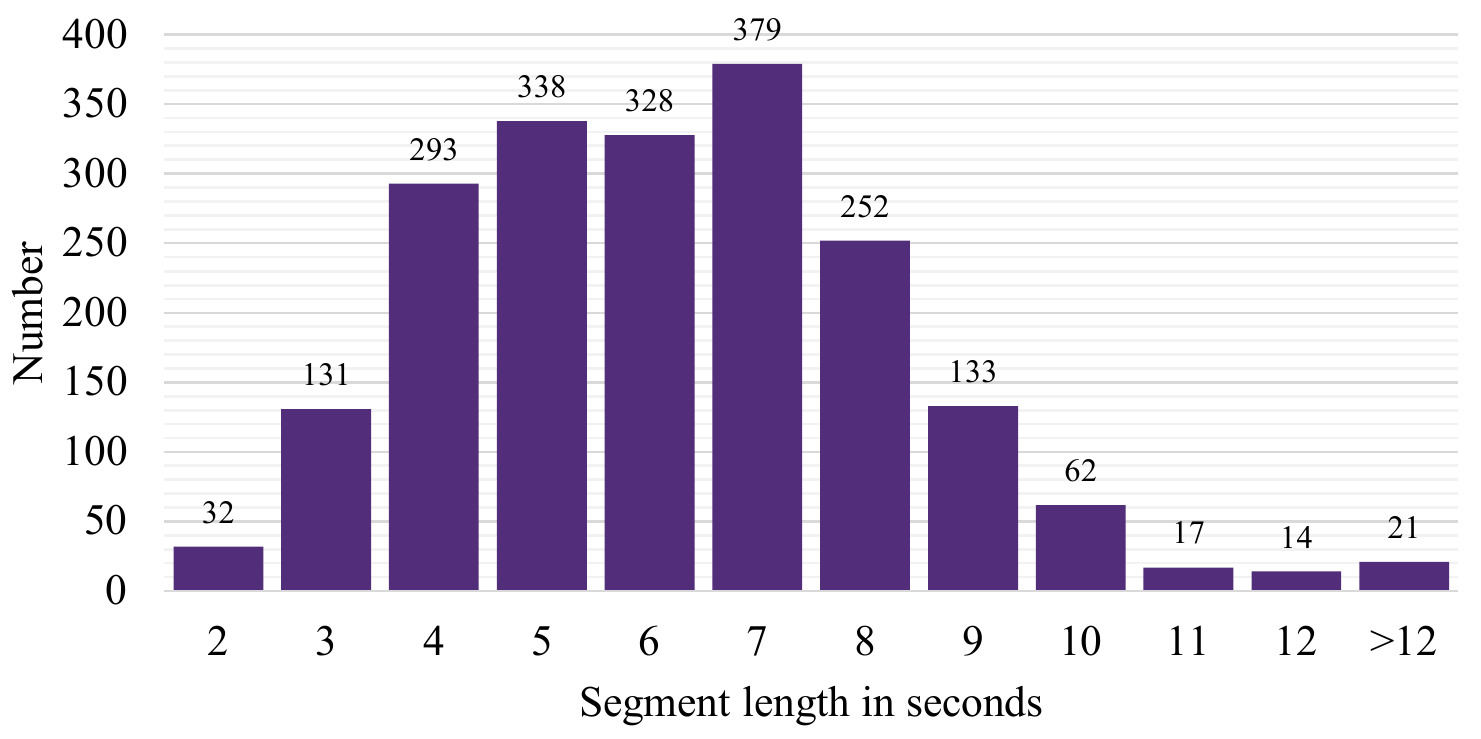}
    \captionsetup{font={scriptsize}}
    \caption{Distribution of temporal segments.}
    \label{fig_tpami:segments}
\end{figure}

%% file: fig/wordcloud.tex
\begin{figure}[t]
    \centering
    \includegraphics[width=0.9\linewidth]{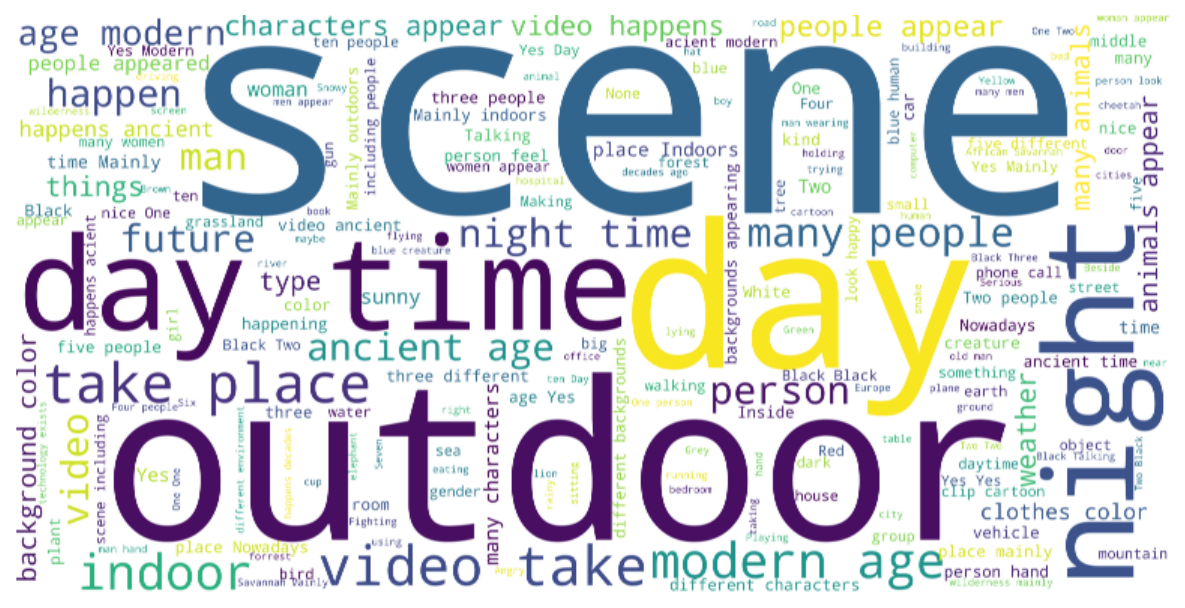}
    \caption{{Word cloud} of the answer set in MovieChat-1K.}
    \label{fig:wordcloud}
\end{figure}

%% file: fig/caption_length.tex
\begin{figure}[t]
    \centering
    \includegraphics[width=0.6\linewidth]{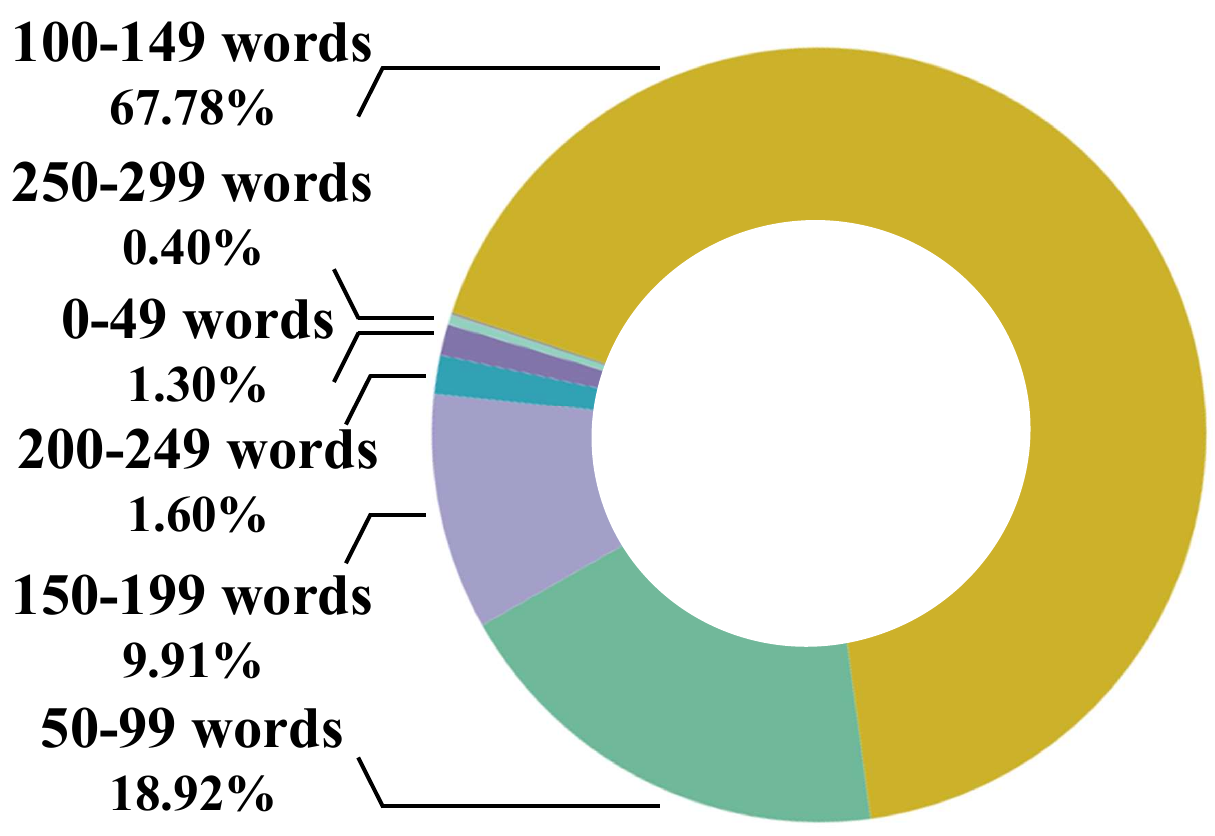}
    \caption{{Distribution of caption length.}}
    \label{fig:caption_length}
\end{figure}

%% file: fig/caption_wordcloud.tex
\begin{figure}[t]
    \centering
    \includegraphics[width=1\linewidth]{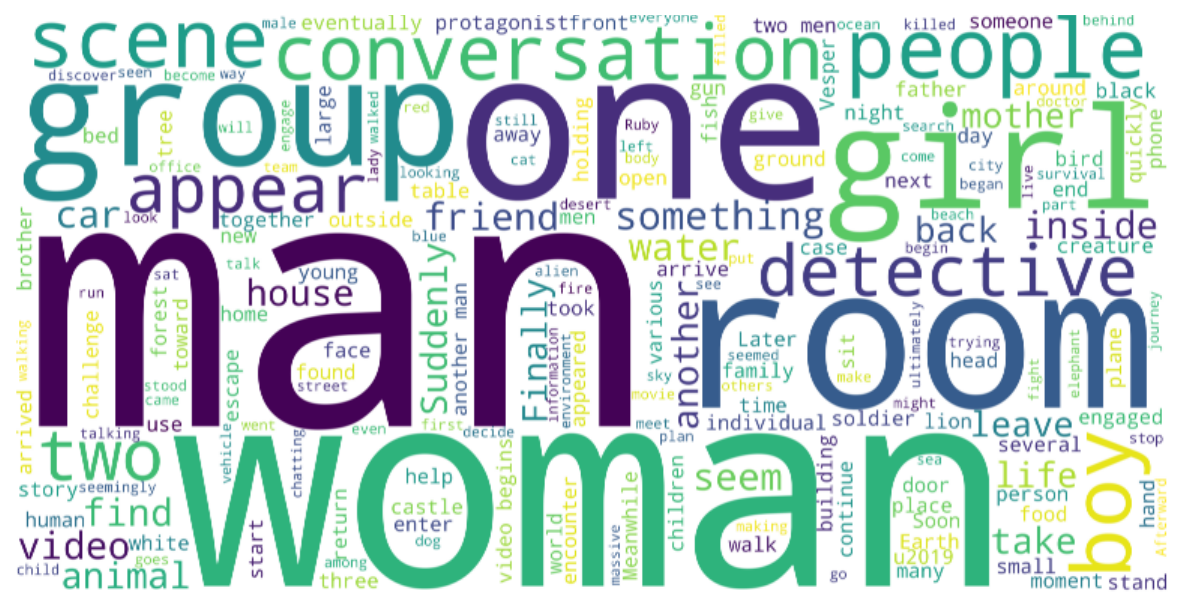}
    \caption{{Word cloud} of the caption set in MovieChat-1K.}
    \label{fig:caption_wordcloud}
\end{figure}

%% file: text_tpami/5_experiment.tex
\input{tab_tpami/short}
\input{tab_tpami/nextqa}

\input{tab_tpami/short_varies}
\input{tab_tpami/seed_bench}

\section{Experiments}

We conduct quantitative and qualitative evaluations of our complete MovieChat+ compared to previous methods and the original MovieChat. Additionally, we perform ablation studies to investigate MovieChat+.

\subsection{Quantitative Evaluation}

\label{exp:quantitative}

\subsubsection{Short Video Question-answering} We use several widely used open-ended datasets: MSVD-QA~\cite{xu2017video}, MSRVTT-QA~\cite{xu2016msr-vtt}, ActivityNet-QA~\cite{yu2019activitynet}, and NExT-QA~\cite{xiao2021next} for short video question-answering tasks. The evaluation process is under the assistance of LLM with the default hyper-parameter settings. The accuracy and relative scores on a scale of $0$ to $5$ are reported. Compared to previous methods~\cite{maaz2023video,li2023videochat,zhang2023llama, zhang2023video, yang2022zero}, MovieChat achieves comparable performance even it is not specifically designed for short video question-answering tasks, as shown in Tab.~\ref{tab_tpami:short}. 

We also report the results of our zero-shot evaluation on the test split of the NExT-QA~\cite{xiao2021next} benchmark in Tab.~\ref{tab_tpami:nextqa}. NExT-QA divides its questions into three categories: Causal (C), Temporal (T), and Description (D). Compared to prior work, our approach achieves higher accuracy across all aspects, demonstrating its effectiveness at understanding temporal context with the question-aware consolidation.


Following~\cite{maaz2023video}, we employ GPT-assisted evaluation to conduct a more comprehensive comparison of the text generation performance between our appraoch and previous methods~\cite{maaz2023video,li2023videochat,zhang2023llama, zhang2023video} on processed ActivityNet-QA~\cite{yu2019activitynet}. The evaluation pipeline covers crucial metrics (including \textit{Correctness of Information}, \textit{Detailed Orientation}, \textit{Contextual Understanding}, \textit{Temporal Understanding} and \textit{Consistency}) and assigns relative scores to the generated predictions on a scale of 0-5. We present the results of the generation performance evaluation in Tab.~\ref{tab_tpami:short_varies}. The results reveal its competitive performance across all key aspects compared to previous methods. 
It should be noted that in comparison with MovieChat, MovieChat+ does not exhibit substantial enhancements in both question-answering accuracy and generative performance when evaluated on short video datasets.
This is because the content of short videos is often closely related to the questions, which is an extreme case for our proposed frame filtering strategy. Similar to the original MovieChat, all video frames are merged with nearly equal consideration.


We further evaluate our approach on the task of action recognition on Seed-Bench~\cite{li2023seed}
to study the effect of MovieChat for short-term temporal understanding tasks. In contrast to the longer setting in the procedure understanding task, the
videos in this task generally have duration of around 10 seconds. 
As shown in Fig.~\ref{tab_tpami:seed}, we compile the number of frames fed into the LLM decoder for different models, along with the corresponding performance of procedure understanding and action recognition. As the input frames increasing, our question-aware approach yields greater benefits in both the procedure understanding task and the action recognition task. These results suggest that the question-aware merge strategy, which filters more related context for reasoning about spatial-temporal relationships between video segments, may be crucial for fine-grained action understanding.
However, when feeding the same number of frames into the LLM decoder, MovieChat shows minimal improvement in procedure understanding. We speculate that when dealing with a limited number of sampling frames, the inclusion of compressed frames containing redundant information could potentially hinder the model's ability to comprehend procedural sequences.

\subsubsection{Long Video Question-answering} We evaluate the long video question-answering performance of MovieChat with our proposed MovieChat-1K. We split 1,000 videos into training set~(800), test set~(100), validation set~(100) and only use test set for final performance evaluation. We select two non-LLM based video understanding models~(\eg GIT~\cite{wang2022git}, and mPLUG-2~\cite{xu2023mplug}) and three recent LLM-based video understanding models~(\eg Video Chat~\cite{li2023videochat}, Video LLaMA~\cite{zhang2023video}, and Video-ChatGPT~\cite{maaz2023video}) as the baselines. Yet, none of those methods can support such long video~($\textgreater 10$K frames). Therefore, to accommodate their length limitations in global questions, we uniformly sample from the original video up to the maximum frame count which can be officially supported by each individual model. For breakpoint questions, we extend half of the maximum frame count before and after the breakpoint ( placing the breakpoint at the center frame). 

\input{tab_tpami/long}

\input{tab_tpami/long_varies}

\input{tab_tpami/break_varies}

\input{tab/type_result_global}

\input{tab/type_result_break}

To enhance the robustness of the results, we simultaneously employ GPT-3.5~\cite{gpt3.5} and Claude~\cite{examplewebpage} as LLM assistants, with the additional support of human blind rating. We observe a discrepancy between the accuracy and relative score generated by the previously LLM-assisted evaluation method~\cite{maaz2023video} for video question-answering tasks. However, merely adjusting the prompt for the LLM cannot effectively address this issue. Therefore, after obtaining the accuracy and score from the LLM-assisted evaluation method, we implement manual filtering to remove results with inconsistent values, thus improving the reliability of our outcomes.

As shown in Tab.~\ref{tab_tpami:long}, compared to previous methods, MovieChat reads more video frames. In both global mode and breakpoint mode, our method maintains a performance gain in terms of the average accuracy and score provided by LLM assistants and human blind rating. Compared with MovieChat, our method significantly improves accuracy in the global mode, which fully demonstrates the effectiveness of our question-aware consolidation strategy.

\input{tab_tpami/ego}
\input{tab_tpami/videollama_score}

\input{tab_tpami/videollama_5}

\input{tab_tpami/memory}


We further compare the quality of answers generated by MovieChat and previous methods~\cite{maaz2023video,li2023videochat,zhang2023video, wang2022git, xu2023mplug} in long video question-answering on MovieChat-1K. As shown in Tab.~\ref{tab_tpami:long_varies} and Tab.~\ref{tab_tpami:break_varies}, with the average score provided by GPT-3.5~\cite{gpt3.5}, Claude~\cite{examplewebpage} and human bling rating, our complete method, MovieChat+, continues to generate higher-quality answers even as the video contents become more extensive, significantly outperforming the initial and simpler version of MovieChat.


MovieChat-1K contains question-answer pairs of varies types. To better assess the performance of our approach, we conduct evaluations on the long video question answering task using various types of questions. We roughly categorize the question types into multiple-choice questions and open-ended questions. With the average results of GPT-3.5~\cite{gpt3.5}, Claude~\cite{examplewebpage} and human blind rating, Tab.~\ref{tab:type_result_global} and Tab.~\ref{tab:type_result_break} respectively present the accuracy and scores of MovieChat and the baseline across different question categories in both global mode and breakpoint mode. In various research conditions, our approach consistently outperforms the baselines in both open-ended and true-false questions.

\subsubsection{Question-answering on Other Long Video Dataset}
We further report zero-shot question-answering results for MovieChat on another commonly used long-form video dataset, EgoSchema~\cite{mangalam2023egoschema} in Tab.~\ref{tab_tpami:ego}. 
EgoSchema~\cite{mangalam2023egoschema} is a diagnostic
benchmark for evaluating long video understanding capabilities of advancing systems, featuring over 5000 human-curated multiple-choice question-answer pairs based on more than 250 hours of real-world video data.
Prior works~\cite{pezeshkpour2023large, zheng2023large, alzahrani2024benchmarks} have demonstrated that the sequence in which options are presented can significantly influence the outcomes of tasks involving multiple choices. To mitigate this effect, we provide MovieChat with questions in EgoSchema~\cite{mangalam2023egoschema}exclusively. Subsequently, we employ LangChain~\cite{langchain} to assess the similarity between the responses of MovieChat and the provided options. We then select the option that most closely aligns with our anticipated answer as our decision.
MovieChat+ produces significantly superior result than its initial version and other leading non-LLM based and LLM-based methods.

\subsection{Ablation Study}

\subsubsection{Short-term and Long-term Memory Buffers} 
As our approach incorporates a memory mechanism including short-term memory and long-term memory, it is imperative to evaluate how the proposed memory mechanism influences the performance. Tab.~\ref{tab_tpami:videollama_score} and Tab.~\ref{tab_tpami:videollama_5} provide the memory-dependent performance of our approach for long video question-answering and generative tasks with the average results of GPT-3.5~\cite{gpt3.5}, Claude~\cite{examplewebpage}, and human blind rating. MovieChat with the memory mechanism significantly outperforms the memory-independent variant, which signifies the importance of memory mechanisms.

We further consider the following approaches to evaluate our memory consolidation strategy:

\noindent\textbf{No memory.} Due to memory constraints, we uniformly sample 16 frames from all frames, concatenate all visual tokens, and feed them into the decoder.

\noindent\textbf{Spatial- or temporal-pooling.} We pool the visual features, along either the spatial or temporal dimensions to reduce
the number of tokens fed to the LLM decoder.

\noindent \textbf{EMA.} Following~\cite{zhou2024streaming}, we use an exponential moving average of frame features at each time step. 

Tab.~\ref{tab_tpami:memory} compares the results of the different memory modules. For $T=16$, where we can feed all the tokens from the vision backbone into the decoder, “No Memory” performs the best because it uses the most tokens. However, it is impractical to use “No Memory” for $T \textgreater 16$ due to its computational cost. With more frames, naively pooling along the spatial- or temporal-dimensions actually performs worse. This is likely because we are averaging out information over longer temporal durations, and thus losing the important details required for more detailed localization or captioning. Our method, on the other hand, leverages more frames to improve performance by retaining diverse features within the memory.




\subsubsection{Large Language Models Ablations}
\input{tab_tpami/llml_score}

\input{fig_tpami/ablation}

Most previous video understanding methods~\cite{maaz2023video,li2023videochat,zhang2023llama, zhang2023video} primarily employed LLama~\cite{touvron2023llama} and its variants~\cite{stablevicuna-github} as text decoders. With the average results of GPT-3.5~\cite{gpt3.5}, Claude~\cite{examplewebpage} and human blind rating, Tab.~\ref{tab_tpami:llm_score} and Tab.~\ref{tab_tpami:llm_5} illustrate how the performance of MovieChat changes when using LLama~\cite{touvron2023llama} and LLama2~\cite{touvron2023llama2} as the large language model respectively.

Contrary to our initial hypothesis, the performance of MovieChat with LLama2~\cite{touvron2023llama2} hardly surpasses those of MovieChat with LLama~\cite{touvron2023llama} across various key metrics. The outcome suggests that the advancements incorporated into LLama2~\cite{touvron2023llama2} may not translate to significant improvements. We further investigate a specific example to analyze this phenomenon. As shown in Fig.~\ref{fig:case}, MovieChat with LLama~\cite{touvron2023llama} provides answers that are more aligned with the video content. Surprisingly, MovieChat with LLama2~\cite{touvron2023llama2} offers an approximation of the time required for each step (indicated in \textit{italics}). While its time estimates do not precisely match the actual durations, the proportion of time provided is realistic. Even though LLama2~\cite{touvron2023llama2} cannot obtain specific time information when processing feature-rich video frames, the memory buffer design allows for dense sampling of video frames, enabling LLama2~\cite{touvron2023llama2} to estimate the proportion of time for each scene based on adjacent similar frames. Therefore, we propose that the lower evaluation metric results of MovieChat with LLama2~\cite{touvron2023llama2} compared to MovieChat with LLama~\cite{touvron2023llama} may be attributed to the question-answer pairs in the dataset.

\input{fig/case}

\input{fig/case1}

\subsubsection{Hyper-parameter Ablations} 
We perform a series of hyper-parameter ablations based on the MovieChat-1K dataset to better understand our approach. Fig.~\ref{fig_tpami:ablation} shows the performance when ablating the length of long-term memory buffer $l_{long}$, the length of short-term memory buffer $l_{short}$, short-term initialization, question-frame similarity $\sigma$, target merging coefficient $\alpha$ and judging relevance basis with the average results of GPT-3.5~\cite{gpt3.5}, Claude~\cite{examplewebpage}, and human blind rating. The performance of MovieChat degrades which shows the validity of our empirically chosen hyper-parameters.

\noindent\textbf{Length of Memory Buffer.} The length of different memory buffers has a combined effect on MovieChat's performance. Since the LLM-based evaluation shows a positive correlation between accuracy and score, we use accuracy to gauge performance. Fig.~\ref{fig_tpami:ablation} (top left and middle) demonstrates that information obtained from the video expands with the growing length of memory buffers. However, this benefit is tempered by a more pronounced loss of fine details, a consequence of maintaining a fixed length for consolidation. Therefore, as the lengths of two memory buffers increase, the performance of our approach exhibits a trend of initially rising and then declining. Thus, we set the length of long/short-term memory to 256 and 16, respectively.

\noindent\textbf{Short-term Initialization.} As shown in Fig.~\ref{fig_tpami:ablation} (top right), using merged tokens for short-term initialization outperforms the last few tokens and uniform sampling. When initializing the next short-term memory with the last few tokens from the previous short-term memory, it is unable to adequately represent the previous information, leading to the final merged tokens being either repetitive or lacking coherence with the previous time step. Uniform sampling faces similar issues, but it manages to capture information with representative frames from the previous time step. 

\input{tab_tpami/llm_5}

\noindent\textbf{Question-frame Similarity Threshold.} Before compressing the short-term memory, we need to assess the relevance of the stored segments to the question by evaluating the similarity between video frames and the question, thereby deciding the degree of compression for the current short-term memory. Fig.~\ref{fig_tpami:ablation} (bottom left) illustrates the outcomes of experimenting with various question-frame similarity thresholds, and the optimal performance is achieved at $\sigma = 0.25$. When the threshold $\sigma$ is low, it is difficult to effectively compress and filter out distracting or redundant information. Conversely, an excessively high threshold $\sigma$ might lead to the over-compression of valuable information.

\noindent\textbf{Target Merging Coefficient.}
We further explore the target merging coefficient $\alpha$ of weakly to strongly related segments. For fairness, all weakly related short-term memories are compressed into 1 frame, allowing us to focus on how the performance of strongly related segments varies with different compression levels. 
Fig.~\ref{fig_tpami:ablation} (bottom middle) shows that increasing merged frames for strongly related segments initially boosts performance but eventually leads to a decline.
We initially assume that less compression of strongly related segments would significantly enhance model performance. Yet, the results hint at a more intricate link between compression intensity and performance. 
To incorporate more long-term memory frames into the pre-trained model, we extend positional encoding with hierarchical decomposition. However, the approach involves balancing extended input lengths with the integrity of positional representations. A direct extension may not be ideal since most training frames are shorter than those in the pre-trained model, making lower positions well-trained for absolute positions, whereas higher positions are less trained, offering only a rough estimate of relative positions. Thus, interpolating lower positions poses a greater risk of disrupting established positional embeddings compared to interpolating higher positions.
When querying the same video with the same question, retaining more merged frames for strongly related segments leads to a noticeably elongated long-term memory, where effective positional encoding becomes challenging, reducing comprehension of long videos. This highlights the necessity of balancing the retention of dense information and the compression for effective long-term video understanding.

\noindent\textbf{Judging Relevance Basis.} Determining the relevance of short-term memory to a question based on similarity can be approached in three ways: comparing the highest similarity within a segment, the lowest, or the average with the question-frame similarity threshold $ \sigma$. According to Fig.~\ref{fig_tpami:ablation} (bottom right), using the minimum or average similarity shows similar performance. However, selecting the maximum similarity as the criterion leads to a performance drop. We believe this is due to the lenient judgment of relevance between the question and frames when choosing the maximum similarity within a segment, which introduces more redundant information. Thus we have elected to utilize the average similarity as the criterion for comparison.


\definecolor{global}{RGB}{21,96,130}
\definecolor{breakpoint}{RGB}{51,0,111}

\subsection{Case Study}

We perform an extensive case study of MovieChat on a variety of open-ended long video~(such as cartoon movie and TV series) including the \parbox[c][8pt][l]{8pt}{\colorbox{breakpoint}{}}breakpoint mode (Q\#1) and the \parbox[c][8pt][l]{8pt}{\colorbox{global}{}}global mode (Q\#2). The evaluation is conducted between our approach and previous methods~\cite{maaz2023video,li2023videochat,zhang2023llama} as shown in Fig.~\ref{fig:case}. For Q\#1 in breakpoint mode, we mark the timestamp when the question is asked. For long videos over $10$K frames, MovieChat is still capable of providing excellent responses to questions regarding both the current moment and the entire video content with less hallucination. We also provide more examples to show the long video scene understanding and temporal understanding ability of MovieChat in Fig.~\ref{fig:case1}, \ref{fig:case2} and \ref{fig:case3}.

%% file: tab_tpami/short.tex
\begin{table}[t]
\centering
\Large
\setlength{\tabcolsep}{7pt}
\renewcommand{\arraystretch}{1.3}
\caption{Quantitative evaluation for short video question answering with GPT-3.5~\cite{gpt3.5}. MovieChat achieves comparable performance even it is not specifically designed for for short video question-answering tasks. The best result is highlighted in bold, and the second best is underlined. Acc. and Sco. stand for accuracy and score respectively.}
\label{tab_tpami:short}
\resizebox{\linewidth}{!}{
\begin{tabular}{@{} l c c c c c c c c @{}}
\toprule
\multirow{1}{*}{\textbf{Method}} & \multicolumn{2}{c}{\textbf{MSVD-QA}} & \multicolumn{2}{c}{\textbf{MSRVTT-QA}} & \multicolumn{2}{c}{\textbf{ActivityNet-QA}} & \multicolumn{2}{c}{\textbf{NExT-QA}} \\
\cline{2-9}
 & \textbf{Acc.} & \textbf{Sco.} & \textbf{Acc.} & \textbf{Sco.} & \textbf{Acc.} & \textbf{Sco.} & \textbf{Acc.} & \textbf{Sco.}\\
\midrule
FrozenBiLM~\cite{yang2022zero} & 2.2 & -- & 16.8 & -- & 24.7 & -- & -- & -- \\
\midrule
Video Chat~\cite{li2023videochat} & 56.3 & 2.8 & 45.0 & 2.5 & 26.5 & 2.2 & \textbf{56.6} & \textbf{3.2} \\
LLaMA Adapter~\cite{zhang2023llama} & 54.9 & 3.1 & 43.8 & \underline{2.7} & 34.2 & \underline{2.7} & -- & --\\
Video LLaMA~\cite{zhang2023video} & 51.6 & 2.5 & 29.6 & 1.8 & 12.4 & 1.1 & -- & --\\
Video-ChatGPT~\cite{maaz2023video} & 64.9 & 3.3 & 49.3 & \textbf{2.8} & 35.2 & \underline{2.7} & 54.6 & \textbf{3.2}\\ 
\midrule
MovieChat & \underline{75.2} & \underline{3.8} &\underline{52.7} & 2.6 & \underline{45.7} & \textbf{3.4} & 49.9 & 2.7\\
MovieChat+ & \textbf{76.5} & \textbf{3.9} &\textbf{53.9} & \underline{2.7} & \textbf{48.1} & \textbf{3.4} & \underline{54.8} & \underline{3.0} \\
\bottomrule
\end{tabular}
}

\end{table}

%% file: tab_tpami/nextqa.tex
\begin{table}[t]
\centering
\setlength{\tabcolsep}{7pt}
\renewcommand{\arraystretch}{0.9}
\caption{Zero-shot evaluation on NExT-QA~\cite{xiao2021next} test split. We observe that our approach performs better than other approaches across most of the different video understanding tasks. $Acc$ stans for the accuracy.The best result is highlighted in bold, and the second best is underlined. }
\label{tab_tpami:nextqa}
\resizebox{\linewidth}{!}{
\begin{tabular}{@{} l c c c c @{}}
\toprule
\scriptsize \textbf{Method} & \scriptsize \textbf{Acc$_{C}$} & \scriptsize \textbf{Acc$_{T}$} & \scriptsize \textbf{Acc$_{D}$} & \scriptsize \textbf{Acc$_{AVG}$}\\
\midrule
\scriptsize Video LLaMA (finetuned) & \scriptsize 27.43 & \scriptsize 32.14 & \scriptsize 32.38 & \scriptsize 29.71 \\
\scriptsize VideoLLama & \scriptsize \underline{31.32} & \scriptsize 35.49 & \scriptsize \underline{42.64} & \scriptsize \underline{34.47}\\
\midrule
\scriptsize MovieChat & \scriptsize 31.12 & \scriptsize \underline{35.80} & \scriptsize 42.49 & \scriptsize 34.43 \\
\scriptsize MovieChat+ & \scriptsize \textbf{32.45} & \scriptsize \textbf{36.03} & \scriptsize \textbf{43.58} & \scriptsize \textbf{35.21}\\
\bottomrule
\end{tabular}
}
\end{table}

%% file: tab_tpami/short_varies.tex
\begin{table}[t]
\centering
\setlength{\tabcolsep}{8.5pt}
\renewcommand{\arraystretch}{0.9}
\caption{Quantitative evaluation for short video generation performance with GPT-3.5~\cite{gpt3.5}. CI stands for correctness of information, DO stands for detail orientation, CU stands for contextual understanding, TU stands for temporal understanding, and CO stands for consistency. The best result is highlighted in bold, and the second best is underlined.}
\label{tab_tpami:short_varies}
\resizebox{\linewidth}{!}{
\begin{tabular}{@{} l c c c c c c c @{}}
\toprule
\scriptsize \textbf{Method} & \scriptsize \textbf{CI} & \scriptsize \textbf{DO} & \scriptsize \textbf{CU} & \scriptsize \textbf{TU} & \scriptsize \textbf{CO}\\
\midrule
 \scriptsize Video Chat~\cite{li2023videochat} &  \scriptsize 2.23& \scriptsize 2.50& \scriptsize 2.53& \scriptsize 1.94& \scriptsize 2.24\\
 \scriptsize LLaMA Adapter~\cite{zhang2023llama}& \scriptsize 2.03& \scriptsize 2.32& \scriptsize 2.30& \scriptsize 1.98& \scriptsize 2.15\\
 \scriptsize Video LLaMA~\cite{zhang2023video} & \scriptsize 1.96& \scriptsize 2.18& \scriptsize 2.16& \scriptsize 1.82& \scriptsize 1.79\\
 \scriptsize Video-ChatGPT~\cite{maaz2023video}& \scriptsize 2.40& \scriptsize 2.52& \scriptsize 2.62& \scriptsize 1.98& \scriptsize 2.37\\
\midrule
\scriptsize MovieChat& \scriptsize \underline{2.76}& \scriptsize \underline{2.93}& \scriptsize \underline{3.01}& \scriptsize \underline{2.24}& \scriptsize \underline{2.42} \\
\scriptsize MovieChat+ &  \textbf{2.87}& \scriptsize \textbf{2.95}& \scriptsize \textbf{3.10}& \scriptsize \textbf{2.25}& \scriptsize \textbf{2.50} \\
\bottomrule
\end{tabular}
}
\end{table}

%% file: tab_tpami/seed_bench.tex
\begin{table}[t]
\centering
\setlength{\tabcolsep}{5pt}
\renewcommand{\arraystretch}{0.8}
\caption{Zero-shot video question answering on Seed-Bench~\cite{li2023seed}. The ``Frames'' column lists the frames input into the LLM decoder. The best result is highlighted in bold, and the second best is underlined. }
\label{tab_tpami:seed}
\resizebox{\linewidth}{!}{
\begin{tabular}{@{} l c c c@{}}
\toprule
\scriptsize \textbf{Method} & \scriptsize \textbf{\# Frames} & \scriptsize \textbf{Procedure} & \scriptsize \textbf{Action}\\
 & & \scriptsize \textbf{Understanding} & \scriptsize \textbf{Recognition} \\
\midrule
\scriptsize Video Chat~\cite{li2023videochat} & \scriptsize 32 & \scriptsize 27.27 & \scriptsize 34.89\\
\scriptsize Video LLaMA~\cite{zhang2023video} & \scriptsize 32 & \scriptsize 25.42 & \scriptsize 35.52 \\
\scriptsize Video-ChatGPT~\cite{maaz2023video} & \scriptsize 32 & \scriptsize 21.14& \scriptsize 27.59\\
\midrule
\multirow{2}{*}{\scriptsize MovieChat} & \scriptsize 32 & \scriptsize 26.76 & \scriptsize 34.37 \\
& \scriptsize 256 & \scriptsize \underline{29.82} & \scriptsize \underline{40.11} \\
\multirow{2}{*}{\scriptsize MovieChat+} & \scriptsize 32 & \scriptsize 27.35 & \scriptsize 36.33 \\
 & \scriptsize 256 & \scriptsize \textbf{31.04} & \scriptsize \textbf{42.45} \\
\bottomrule
\end{tabular}
}
\end{table}

%% file: tab_tpami/long.tex

\begin{table}[t]
\centering
\setlength{\tabcolsep}{5pt}
\renewcommand{\arraystretch}{1.3}
\caption{Quantitative evaluation for long video question answering on MovieChat-1K test set in global mode with the average of GPT-3.5~\cite{gpt3.5}, Claude~\cite{examplewebpage} and human bling rating. "\# Frames" indicates the count of video frames read by the models. The best result is highlighted in bold, and the second best is underlined. Acc. and Sco. stand for accuracy and score respectively.}
\label{tab_tpami:long}
\resizebox{\linewidth}{!}{
\begin{tabular}{@{} l c c c c c c @{}}
\toprule
\multirow{2}{*}{\textbf{Method}} & \multirow{2}{*}{\textbf{Text Decoder}} & \multirow{2}{*}{\textbf{\# Frames}} & \multicolumn{2}{c}{\textbf{Global Mode}} & \multicolumn{2}{c}{\textbf{Breakpoint Mode}} \\
\cline{4-7}
 & &  & \textbf{Acc.} & \textbf{Sco.} & \textbf{Acc.} & \textbf{Sco.} \\
\midrule
GIT~\cite{wang2022git} & non-LLM based & 6 & 28.8 & 1.83 & 29.2 & 1.98 \\
mPLUG-2~\cite{xu2023mplug} &  non-LLM based & 8 &  31.7 & 2.13 &  30.8 & 1.83 \\
\midrule
Video Chat~\cite{li2023videochat}& LLM based & 32 & 57.8 & 3.00 & 46.1 & 2.29 \\
Video LLaMA~\cite{zhang2023video}& LLM based & 32 & 51.7 & 2.67 & 39.1 & 2.04 \\
Video-ChatGPT~\cite{maaz2023video}& LLM based & 100 & 47.6 & 2.55 & 48.0 & 2.45 \\ 
\midrule
MovieChat & LLM based & 2048 & \underline{62.3} & \underline{3.23} & \underline{48.3} & \underline{2.57}\\
MovieChat+ & LLM based & 2048 & \textbf{71.2} & \textbf{3.51} & \textbf{49.6} & \textbf{2.62}\\
\bottomrule
\end{tabular}
}
\end{table}

%% file: tab_tpami/long_varies.tex
\begin{table}[t]
\centering
\setlength{\tabcolsep}{6pt}
\renewcommand{\arraystretch}{0.9}
\caption{Quantitative evaluation for long video generation performance in global mode with the average of GPT-3.5~\cite{gpt3.5}, Claude~\cite{examplewebpage} and human blind rating. CI stands for correctness of information, DO stands for detail orientation, CU stands for contextual understanding, TU stands for temporal understanding, and CO stands for consistency. The best result is in bold, and the second best is underlined.}
\label{tab_tpami:long_varies}
\resizebox{\linewidth}{!}{
\begin{tabular}{@{} l c c c c c c c @{}}
\toprule
\scriptsize \textbf{Method} & \scriptsize \textbf{Text Decoder} & \scriptsize \textbf{CI} & \scriptsize \textbf{DO} & \scriptsize \textbf{CU} & \scriptsize \textbf{TU} & \scriptsize \textbf{CO}\\
\midrule
GIT~\cite{wang2022git} & \scriptsize non-LLM based & 1.65 & 1.74 & 2.00 & 1.88 & 2.16 \\
mPLUG-2~\cite{xu2023mplug} & \scriptsize non-LLM based &  1.82 & 1.93 & 2.24 & 1.92 & 2.35 \\
\midrule
\scriptsize Video Chat~\cite{li2023videochat} & \scriptsize LLM based & \scriptsize 3.04 & \scriptsize 2.75 & \scriptsize 3.09 & \scriptsize 3.00& \scriptsize  3.21\\
\scriptsize Video LLaMA~\cite{zhang2023video} & \scriptsize LLM based & \scriptsize 2.75 & \scriptsize 2.24& \scriptsize 2.83& \scriptsize 2.62& \scriptsize 2.97\\
\scriptsize Video-ChatGPT~\cite{maaz2023video}& \scriptsize LLM based & \scriptsize 2.37& \scriptsize 2.30& \scriptsize 2.58& \scriptsize 2.49& \scriptsize 2.69\\
\midrule
\scriptsize MovieChat & \scriptsize LLM based & \scriptsize \underline{3.11}& \scriptsize \underline{2.93} & \scriptsize \underline{3.24}& \scriptsize \underline{3.17}& \scriptsize \underline{3.25} \\
\scriptsize MovieChat+ & \scriptsize LLM based & \scriptsize \textbf{3.28}& \scriptsize \textbf{3.04} & \scriptsize \textbf{3.47}& \scriptsize \textbf{3.22}& \scriptsize \textbf{3.41} \\
\bottomrule
\end{tabular}
}
\end{table}

%% file: tab_tpami/break_varies.tex
\begin{table}[t]
\centering
\setlength{\tabcolsep}{9pt}
\renewcommand{\arraystretch}{0.8}
\caption{Quantitative evaluation for long video generation performance in breakpoint mode with the average of GPT-3.5~\cite{gpt3.5}, Claude~\cite{examplewebpage} and human blind rating. CI stands for correctness of information, DO stands for detail orientation, CU stands for contextual understanding, TU stands for temporal understanding, and CO stands for consistency. The best result is highlighted in bold, and the second best is underlined.}
\label{tab_tpami:break_varies}
\resizebox{\linewidth}{!}{
\begin{tabular}{@{} l c c c c c c @{}}
\toprule
\scriptsize \textbf{Method} & \scriptsize \textbf{Text Decoder} & \scriptsize \textbf{CI} & \scriptsize \textbf{DO} & \scriptsize \textbf{CU} & \scriptsize \textbf{TU} & \scriptsize \textbf{CO}\\
\midrule
GIT~\cite{wang2022git} & \scriptsize non-LLM based & 1.44 & 1.48 & 1.96 & 1.77 & 2.01 \\
mPLUG-2~\cite{xu2023mplug} & \scriptsize non-LLM based & 1.53 & 1.62 & 1.91 & 1.56 & 2.14 \\
\midrule
\scriptsize Video Chat~\cite{li2023videochat} & \scriptsize LLM based & \scriptsize 2.42 & \scriptsize 2.51 & \scriptsize 2.81 & \scriptsize 2.10 & \scriptsize 2.78\\
\scriptsize Video LLaMA~\cite{zhang2023video} & \scriptsize LLM based & \scriptsize 2.04 & \scriptsize 2.29& \scriptsize 2.63& \scriptsize 2.00& \scriptsize 2.87\\
\scriptsize Video-ChatGPT~\cite{maaz2023video} & \scriptsize LLM based & \scriptsize 2.62& \scriptsize \textbf{2.65}& \scriptsize \underline{2.86}& \scriptsize 2.32 & \scriptsize 2.96\\
\midrule
\scriptsize MovieChat& \scriptsize LLM based & \scriptsize \underline{2.64}& \scriptsize 2.60& \scriptsize \textbf{2.87}& \scriptsize \underline{2.49}& \scriptsize \textbf{3.08} \\
\scriptsize MovieChat+ & \scriptsize non-LLM based & \scriptsize \textbf{2.65}& \scriptsize \underline{2.62}& \scriptsize \textbf{2.87}& \scriptsize \textbf{2.51}& \scriptsize \underline{3.07} \\
\bottomrule
\end{tabular}
}
\end{table}

%% file: tab/type_result_global.tex
\begin{table}[t]
\centering
\setlength{\tabcolsep}{6pt}
\renewcommand{\arraystretch}{1.0}
\caption{Quantitative evaluation for long video different types question answering in global mode. The best result is highlighted in bold, and the second best is underlined. Acc. and Sco. stand for accuracy and score respectively.}
\label{tab:type_result_global}
\resizebox{\linewidth}{!}{
\begin{tabular}{@{} l c c c c c c @{}}
\toprule
\multirow{2}{*}{\vspace{-1ex} \textbf{Method}} & \multicolumn{2}{c}{ \textbf{Total}} & \multicolumn{2}{c}{ \textbf{Multi-choice}} & \multicolumn{2}{c}{ \textbf{Open-ended}} \\
\cline{2-7}
 &  \textbf{Acc.} &  \textbf{Sco.} &  \textbf{Acc.} &  \textbf{Sco.} &  \textbf{Acc.} &  \textbf{Sco.}\\
\midrule
Video Chat~\cite{li2023videochat}& 61.0& 3.34 & 74.8 & 3.83 & 56.4 & 3.02 \\
Video LLaMA~\cite{zhang2023video}& 51.4&3.10 & 78.3 & 3.58 & 38.8 & 2.67 \\
Video-ChatGPT~\cite{maaz2023video}& 44.2& 2.71 & 52.5 & 3.16 & 37.7 & 2.54 \\ 
\midrule
MovieChat & \underline{62.3} & \underline{3.81} &\underline{80.9} & \underline{4.02} & \underline{57.5} & \underline{3.74}\\
MovieChat+ & \textbf{71.2} & \textbf{3.51} & \textbf{81.4} & \textbf{4.03} & \textbf{60.1} & \textbf{3.79} \\
\bottomrule
\end{tabular}
}

\end{table}

%% file: tab/type_result_break.tex
\begin{table}[t]
\centering
\setlength{\tabcolsep}{6pt}
\renewcommand{\arraystretch}{1.0}
\caption{Quantitative evaluation for long video different types question answering in breakpoint mode. The best result is highlighted in bold, and the second best is underlined. Acc. and Sco. stand for accuracy and score respectively.}
\label{tab:type_result_break}
\resizebox{\linewidth}{!}{
\begin{tabular}{@{} l c c c c c c @{}}
\toprule
\multirow{2}{*}{\vspace{-1ex} \textbf{Method}} & \multicolumn{2}{c}{ \textbf{Total}} & \multicolumn{2}{c}{ \textbf{Multi-choice}} & \multicolumn{2}{c}{ \textbf{Open-ended}} \\
\cline{2-7}
 &  \textbf{Acc} &  \textbf{Sco.} &  \textbf{Acc} &  \textbf{Sco.} &  \textbf{Acc} &  \textbf{Sco.}\\
\midrule
Video Chat~\cite{li2023videochat}& 48.3& 2.43 & \underline{62.4} & 3.46 & 44.5 & 2.19 \\
Video LLaMA~\cite{zhang2023video}& 38.2& 2.33 & 57.3 & 2.39 & 33.1 & 2.31 \\
Video-ChatGPT~\cite{maaz2023video}& 49.8 & 2.71 & 58.3 & 3.05 & 47.5 & 2.37 \\ 
\midrule
MovieChat & \underline{48.3} & \underline{2.57} & \underline{62.4} & \underline{3.65} & \underline{46.7} & \underline{2.70}\\
MovieChat+ & \textbf{49.6} & \textbf{2.62} & \textbf{64.0} & \textbf{3.79} & \textbf{47.7} & \textbf{2.73}\\
\bottomrule
\end{tabular}
}

\end{table}

%% file: tab_tpami/ego.tex

\begin{table}[h]
\centering
\scriptsize
\setlength{\tabcolsep}{8pt}
\renewcommand{\arraystretch}{0.5}
\captionsetup{font={scriptsize}}
\caption{Zero-shot QA Evaluation on EgoSchema. Acc. stands for the accuracy. The best result is highlighted in bold, and the second best is underlined.}
\label{tab_tpami:ego}
\resizebox{\linewidth}{!}{
\begin{tabular}{@{} l c c @{}}
\toprule
\tiny \textbf{Model}
 & \tiny \textbf{Text Decoder}  & \tiny \textbf{Acc.} \\
\midrule
\multicolumn{2}{c}{\tiny Choosing the correct answer uniformly at random} & \tiny  20.0\%  \\
\midrule
\tiny FrozenBiLM~\cite{yang2022zero} & \tiny non-LLM based & \tiny 26.9 \\
\midrule
\tiny Video Chat~\cite{li2023videochat} & \tiny LLM based & \tiny 47.5 \\
\tiny MovieChat  & \tiny LLM based & \tiny 53.5 \\
\tiny MovieChat+ & \tiny LLM based & \tiny \underline{56.4} \\
\midrule
\multicolumn{2}{c}{\tiny Human Performance} & \tiny \textbf{76.2\%} \\
\bottomrule
\end{tabular}
}
\end{table}

%% file: tab_tpami/videollama_score.tex
\begin{table}[t]
\centering
\setlength{\tabcolsep}{12pt}
\renewcommand{\arraystretch}{0.7}
\caption{Ablation study on how memory mechanism (MM) affects the long video question answering. The best result is in bold.}
\label{tab_tpami:videollama_score}
\resizebox{\linewidth}{!}{
\begin{tabular}{@{} c c c c c @{}}
\toprule
\multirow{2}{*}{\textbf{\scriptsize Method}} & \multicolumn{2}{c}{\textbf{\scriptsize Global Mode}} & \multicolumn{2}{c}{\textbf{\scriptsize Breakpoint Mode}} \\
\cline{2-5}
 & \scriptsize \textbf{Accuracy} & \scriptsize \textbf{Score} & \scriptsize \textbf{Accuracy} &  \scriptsize \textbf{Score}\\
\midrule
 \rule{0pt}{5pt} \scriptsize w/o MM &   \scriptsize 51.4&  \scriptsize 3.10&  \scriptsize 38.2&  \scriptsize 2.31 \\
  \rule{0pt}{5pt} \scriptsize base&  \scriptsize \textbf{67.8}&  \scriptsize \textbf{3.81}&  \scriptsize \textbf{50.4}& \scriptsize \textbf{2.96} \\
\bottomrule
\end{tabular}
}

\end{table}

%% file: tab_tpami/videollama_5.tex
\begin{table}[t]
\centering
\setlength{\tabcolsep}{5pt}
\renewcommand{\arraystretch}{1.3}
\caption{Ablation study on how memory mechanism (MM) affects the long video generative performance. CI stands for correctness of information, DO stands for detail orientation, CU stands for contextual understanding, TU stands for temporal understanding, and CO stands for consistency. The best result is in bold.}
\label{tab_tpami:videollama_5}
\resizebox{\linewidth}{!}{
\begin{tabular}{@{} c c c c c c c c c c c @{}}
\toprule
\multirow{2}{*}{\textbf{Method}} & \multicolumn{5}{c}{\textbf{Global Mode}} & \multicolumn{5}{c}{\textbf{Breakpoint Mode}} \\
\cline{2-11}
& \textbf{CI} & \textbf{DO} & \textbf{CU} & \textbf{TU} & \textbf{CO} & \textbf{CI} & \textbf{DO} & \textbf{CU} & \textbf{TU} & \textbf{CO}\\
\midrule
w/o MM &  3.30&  2.53&  3.28&  2.77& 3.42& 2.42& 2.85& 2.87& 2.00& 2.87 \\
base&  \textbf{3.32}&  \textbf{3.28}&  \textbf{3.40}&  \textbf{2.97}& \textbf{3.48}& \textbf{2.97}& \textbf{3.24}& \textbf{3.31}& \textbf{2.70}& \textbf{3.45}\\
\bottomrule
\end{tabular}
}

\vspace{-5pt}
\end{table}

%% file: tab_tpami/memory.tex
\begin{table}[t]
\centering
\setlength{\tabcolsep}{5pt}
\renewcommand{\arraystretch}{1.0}
\caption{Ablation on memory modules. We show CIDEr on ActivityNet~\cite{yu2019activitynet} under different input frames $T$. The second column shows the number of input tokens to the LLM decoder. $N_f=257$ is the number of tokens per-frame. $K=514$ is the number of memory tokens. The best result is highlighted in bold, and the second best is underlined.}
\label{tab_tpami:memory}
\resizebox{\linewidth}{!}{
\begin{tabular}{@{} l c c c c c  @{}}
\toprule
\scriptsize \textbf{Method} & \scriptsize \textbf{\# Tokens} & \scriptsize \textbf{$T=16$} & \scriptsize \textbf{$T=32$} & \scriptsize \textbf{$T=64$} & \scriptsize \textbf{$T=128$} \\
\midrule
\scriptsize No Memory & \scriptsize $T \times N_f$ & \scriptsize \textbf{29.8} & \scriptsize - & \scriptsize - & \scriptsize - \\
\scriptsize Spatial Pooling & \scriptsize $T$ & \scriptsize 27.6 & \scriptsize 27.3 & \scriptsize \underline{27.9} & \scriptsize \underline{27.4} \\
\scriptsize Temporal Pooling & \scriptsize $N_f$ & \scriptsize \underline{29.3} & \scriptsize \underline{28.0} & \scriptsize 26.8 & \scriptsize 25.2 \\
\scriptsize EMA~\cite{zhao2022real} & \scriptsize $N_f$ & \scriptsize 28.2 & \scriptsize 26.3 & \scriptsize 22.0 & \scriptsize 16.3 \\
\midrule
\scriptsize Ours & \scriptsize $K$ & \scriptsize \underline{29.3} & \scriptsize \textbf{29.2} & \scriptsize \textbf{28.9} & \scriptsize \textbf{29.3} \\
\bottomrule
\end{tabular}
}
\end{table}

%% file: tab_tpami/llml_score.tex
\begin{table}[h]
\centering
\setlength{\tabcolsep}{6pt}
\renewcommand{\arraystretch}{0.7}
\caption{Ablation Study on how LLM affects the long video question answering. The best result is highlighted in bold.}
\label{tab_tpami:llm_score}
\resizebox{\linewidth}{!}{
\begin{tabular}{@{} l c c c c @{}}
\toprule
\multirow{2}{*}{
\tiny \textbf{Method}} & \multicolumn{2}{c}{\tiny \textbf{Global Mode}} & \multicolumn{2}{c}{\tiny \textbf{Breakpoint Mode}} \\
\cline{2-5}
 & \tiny \textbf{Accuracy} & \tiny \textbf{Score} & \tiny \textbf{Accuracy} & \tiny \textbf{Score}\\
\midrule
\tiny LLama~\cite{touvron2023llama} & \tiny \textbf{67.8}& \tiny \textbf{3.81}& \tiny \textbf{50.4}& \tiny 2.96 \\
\tiny LLama2~\cite{touvron2023llama2}& \tiny 64.2& \tiny 3.79& \tiny 48.1& \tiny \textbf{2.98} \\
\bottomrule
\end{tabular}
}
\end{table}

%% file: fig_tpami/ablation.tex
\begin{figure*}[t]
  \centering
  \includegraphics[width=\textwidth]{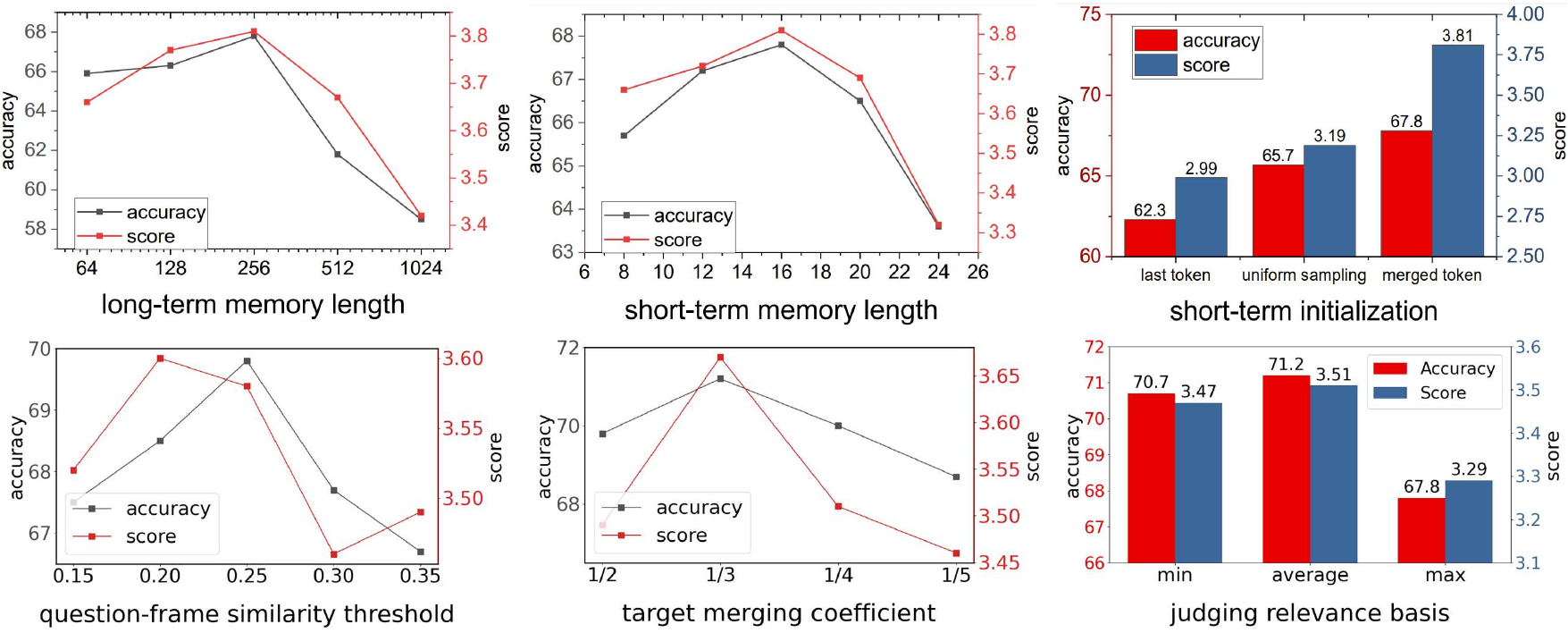}
  \caption{Hyperparameter ablation studies on how length of memory buffer $l_{long}$ and $l_{short}$, short-term initialization, question-frame similarity threshold $\sigma$, target merging coefficient $\alpha$ and judging relevance basis affect the performance of MovieChat+ on long video understanding. }
  \label{fig_tpami:ablation}
\end{figure*}

%% file: fig/case.tex


\begin{figure*}[!]
	\centering
    \includegraphics[width=0.96\linewidth]{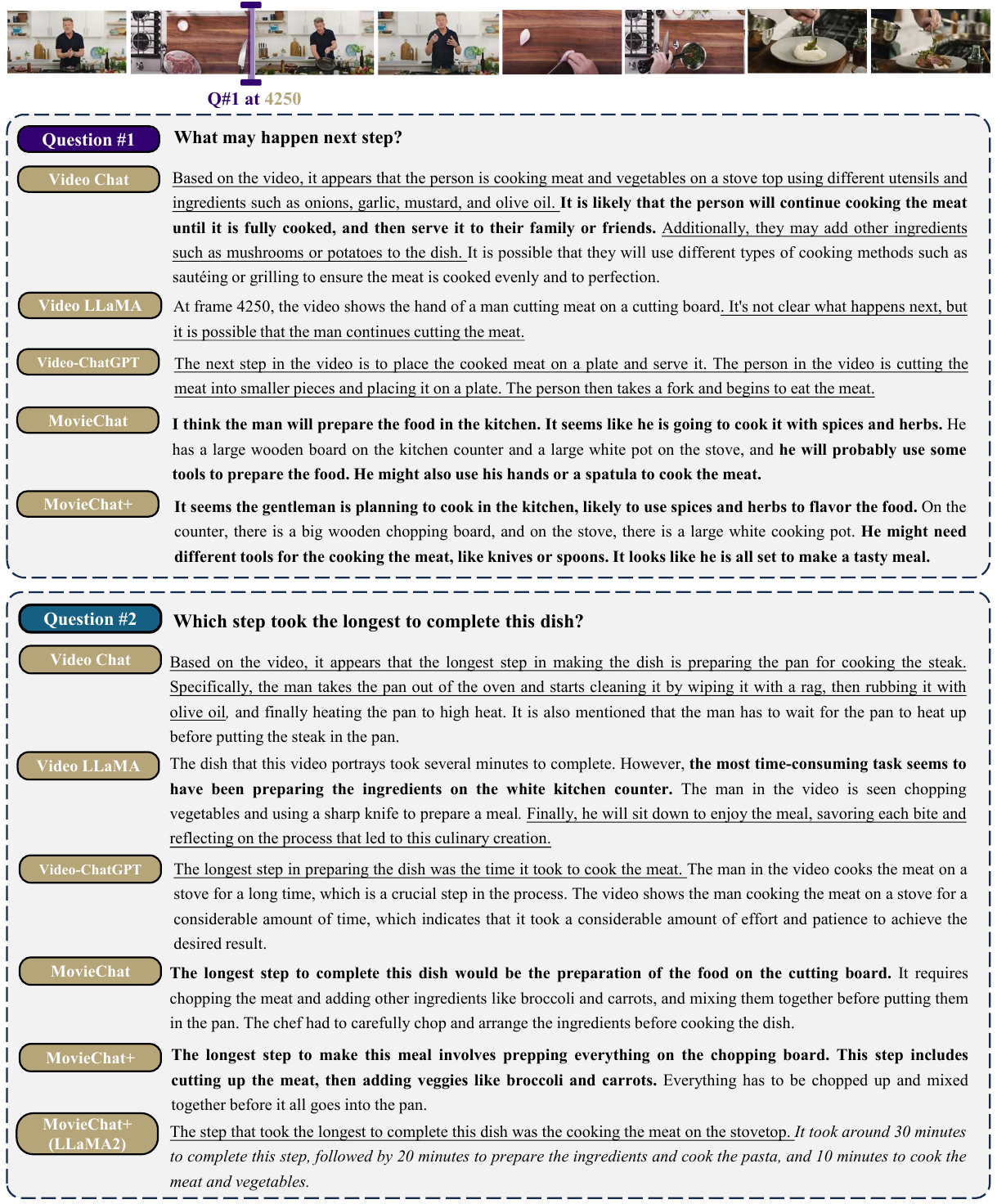}
	\caption{Question and answer about a clip from \textit{YouTube}, which is a tutorial on how to cook steak. The entire instructional process begins with marinating the steak, followed by pan-searing it, preparing side dishes, and ultimately plating the meal. The correct answer is highlighted in bold, the wrong answer or the hallucinating part is underlined, and the answer with an approximation of the time required is indicated in italics.}
 \vspace{20pt}
\label{fig:case}
\end{figure*}

%% file: fig/case1.tex
\begin{figure*}[!t]
	\centering
    \includegraphics[width=\linewidth]{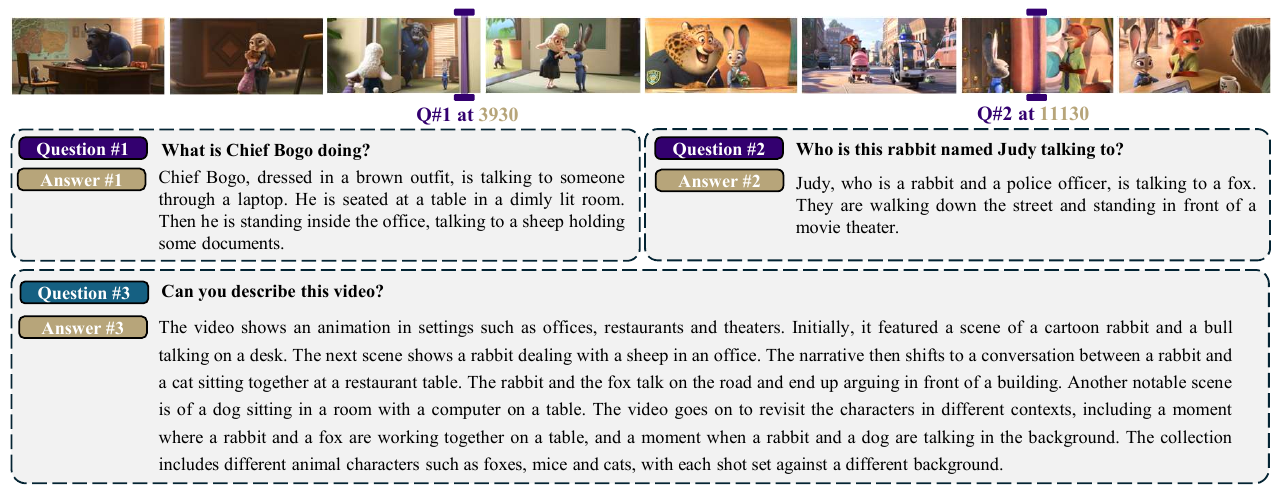}
	\caption{Question and answer about clips from \textit{Zootopia}, a cartoon, which tells the story of a determined police officer rabbit named Judy who pairs up with a cunning fox to uncover a conspiracy about missing animals and develop an unexpected friendship.}
        \label{fig:case1}
    \vspace{15pt}
    \includegraphics[width=\linewidth]{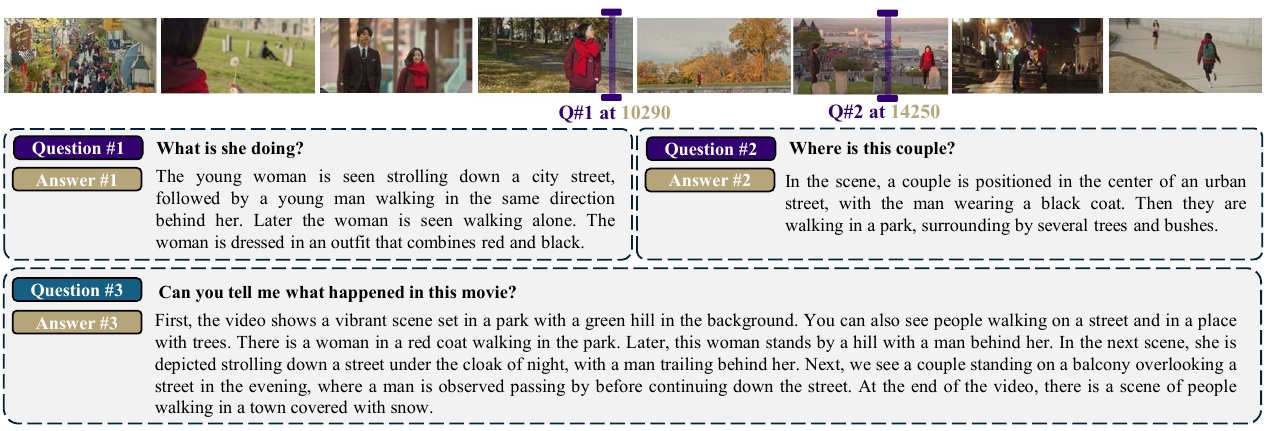}
	\caption{Question and answer about clips from \textit{Goblin}, which tells the story of Kim Shin, an immortal "goblin" who needs to find a human bride to end his endless life but instead meets Ji Eun-tak, a girl fated to die who claims to be the "goblin's bride," leading to a romantic tale unfolding between them.}
        \label{fig:case2}
    \vspace{15pt}
 \includegraphics[width=\linewidth]{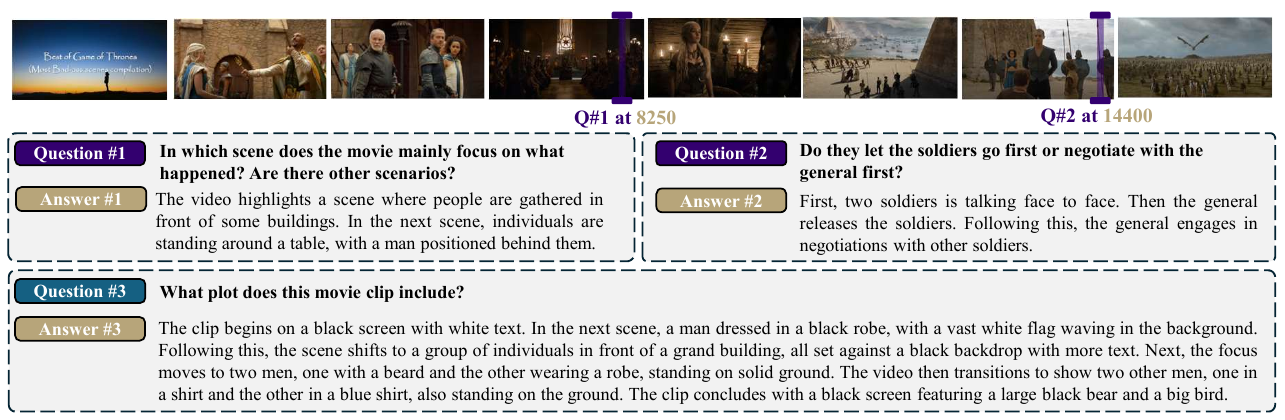}
	\caption{Question and answer about clips from \textit{Game of Thrones}, which tells the epic fantasy tale of power struggles and political intrigue among the Seven Kingdoms, entwined with intricate family relationships, all set against the backdrop of an ancient, mystical threat.}
    \label{fig:case3}

\end{figure*}

%% file: tab_tpami/llm_5.tex
\begin{table}[h]
\centering
\setlength{\tabcolsep}{4pt}
\renewcommand{\arraystretch}{1.3}
\caption{Ablation Study on how the large language model affects the long video generative performance. MM stands for memory mechanism, CI stands for correctness of information, DO stands for detail orientation, CU stands for contextual understanding, TU stands for temporal understanding, and CO stands for consistency. The best result is highlighted in bold.}
\label{tab_tpami:llm_5}
\resizebox{\linewidth}{!}{
\begin{tabular}{@{} l c c c c c c c c c c @{}}
\toprule
\multirow{2}{*}{\textbf{Method}} & \multicolumn{5}{c}{\textbf{Global Mode}} & \multicolumn{5}{c}{\textbf{Breakpoint Mode}} \\
\cline{2-11}
& \textbf{CI} & \textbf{DO} & \textbf{CU} & \textbf{TU} & \textbf{CO} & \textbf{CI} & \textbf{DO} & \textbf{CU} & \textbf{TU} & \textbf{CO}\\
\midrule
LLama~\cite{touvron2023llama}&  \textbf{3.32}&  \textbf{3.28}&  3.40&  \textbf{2.97}& \textbf{3.48}& \textbf{2.97}& \textbf{3.24}& 3.31& \textbf{2.70}& 3.45 \\
LLama2~\cite{touvron2023llama2}&  3.27&  \textbf{3.28}&  \textbf{3.41} &  2.95& 3.45 & 2.96& 3.12& \textbf{3.38}& 2.68& \textbf{3.34}\\
\bottomrule

\end{tabular}
}

\end{table}

%% file: text_tpami/6_conclusion.tex
\section{Limitation}

Although MovieChat has demonstrated impressive abilities in long video understanding, it is still an early-stage prototype and has some limitations, including 1) Limited perception capacities. The performance of our approach is hindered by the pre-trained short video understanding model. 2) Inadequate Time Processing. MovieChat provides only rough estimates of the duration proportions of events within long videos, lacking precision in temporal details.

\section{Conclusion}

Conclusively, we present an innovative video understanding system integrating video foundation models and large language models. By incorporating an enhanced memory mechanism represented by tokens in Transformers, our proposed system, MovieChat overcomes challenges associated with analyzing long videos.
MovieChat achieves state-of-the-art performance in long video understanding, surpassing existing systems, which are limited to handling videos with few frames.
